\documentclass[a4paper,fleqn]{cas-sc}

\usepackage[authoryear]{natbib}
\usepackage{makecell} 
\usepackage{graphicx}
\usepackage{threeparttable}
\usepackage{xcolor}
\usepackage{colortbl}
\usepackage{booktabs}
\usepackage{multirow}
\usepackage[table]{xcolor}
\usepackage{siunitx}
\usepackage{lineno}

\usepackage{setspace}

\begin{document}
\let\printorcid\relax
\let\WriteBookmarks\relax
\def\floatpagepagefraction{1}
\def\textpagefraction{.001}

\shorttitle{}    

\shortauthors{}  

\title [mode = title]{Progressive Scale Convolutional Network for Spatio-Temporal Downscaling of Soil Moisture: A Case Study Over the Tibetan Plateau}

%

\author[1]{Ziyu Zhou}[
          type= author,
          auid=000,
          style=chinese,]
\credit{Methodology, Software, Data curation, Visualization, Writing – review \& editing}
\fnref{equal}

\affiliation[1]{organization={College of Geography and Remote Sensing, Hohai University},
            city={Nanjing},
            postcode={211100}, 
            country={China}}

\author[2]{Keyan Hu}[
  type= author,
  auid=001,
       style=chinese,]
\credit{Methodology,  Formal analysis, Visualization, Writing – original draft}
\fnref{equal}
\fntext[equal]{Ziyu Zhou and Keyan Hu are co-first authors.}
\affiliation[2]{organization={School of Geosciences and Info-Physics, Central South University},
                city={Changsha},
                postcode={410100}, 
                state={Hunan},
                country={China}}

\author[3]{Ling Zhang}[
  type= author,
  auid=005,
       style=chinese,]
\credit{Formal analysis, Writing – review \& editing}
\affiliation[3]{organization={School of Naval Architecture \& Intelligent Manufacturing, Jiangsu Maritime Institute},
            city={Nanjing},
            postcode={211100}, 
            country={China}}

\author[1,4]{Zhaohui Xue}[
              type= author,
              auid=002,
              style=chinese,]
\credit{Conceptualization, Supervision, Methodology, Funding acquisition, Writing – review \& editing}

\affiliation[4]{organization={Key Laboratory of Soil and Water Processes in Watershed, Hohai University},
            city={Nanjing},
            postcode={211100}, 
            country={China}}
\ead{xzh2012@163.com}
\cormark[1]
\cortext[1]{Corresponding author}

\author[5]{Yutian Fang}[
  type= author,
  auid=003,
       style=chinese,]
\credit{Data curation,  Formal analysis, Visualization}
\affiliation[5]{organization={School of Earth Sciences and Engineering, Hohai University},
            city={Nanjing},
            postcode={211100}, 
            country={China}}
\author[5]{Yusha Zheng}[
  type= author,
  auid=004,
       style=chinese,]
\credit{Formal analysis, Visualization, Writing – review \& editing}

\begin{abstract}
Soil moisture (SM) plays a critical role in hydrological and meteorological processes. High-resolution SM can be obtained by combining coarse passive microwave data with fine-scale auxiliary variables. However, the inversion of SM at the temporal scale is hindered by the incompleteness of surface auxiliary factors. To address this issue, first, we introduce validated high temporal resolution ERA5-Land variables into the downscaling process of the low-resolution SMAP SM product. Subsequently, we design a progressive scale convolutional network (PSCNet), at the core of which are two innovative components: a multi-frequency temporal fusion module (MFTF) for capturing temporal dynamics, and a bespoke squeeze-and-excitation (SE) block designed to preserve fine-grained spatial details. Using this approach, we obtained seamless SM products for the Tibetan Plateau (TP) from 2016 to 2018 at 10-km spatial and 3-hour temporal resolution. The experimental results on the TP demonstrated the following: 1) In the satellite product validation, the PSCNet exhibited comparable accuracy and lower error, with a mean R value of 0.881, outperforming other methods. 2) In the in-situ site validation, PSCNet consistently ranked among the top three models for the R metric across all sites, while also showing superior performance in overall error reduction. 3) In the temporal generalization validation, the feasibility of using high-temporal resolution ERA5-Land variables for downscaling was confirmed, as all methods maintained an average relative error within 6\% for the R metric and 2\% for the ubRMSE metric. 4) In the temporal dynamics and visualization validation, PSCNet demonstrated excellent temporal sensitivity and vivid spatial details. Overall, PSCNet provides a promising solution for spatio-temporal downscaling by effectively modeling the intricate spatio-temporal relationships in SM data. 
\end{abstract}




\maketitle

\section{Introduction}
Soil moisture (SM) represents the volumetric water content in unsaturated soil~\citep{seneviratne2010investigating}. It constitutes a fundamental element in hydrology and meteorology~\citep{chahine1992hydrological,mccoll2017global}, regulating evapotranspiration and impacting seepage, runoff, and groundwater flow~\citep{akbar2020partitioning,sadeghi2020global}. It is also a core variable in the terrestrial-atmospheric energy cycle dynamics
~\citep{dobriyal2012review,petropoulos2015surface}, modulating the partitioning of available energy into sensible and latent heat fluxes at the Earth's surface~\citep{sabaghy2018spatially,xia2014evaluation}, thereby influencing the equilibrium of water, energy, and biogeochemical systems~\citep{jung2017compensatory,ochsner2013state,vereecken2022soil}. 

Reliable SM data can be obtained through three primary approaches: in-situ measurements, reanalysis datasets, and satellite remote sensing observations~\citep{dorigo2011new,njoku1996passive,zhang2022dataset}. In-situ measurements of SM provide high local accuracy. However, the spatial variability of climatic conditions and significant heterogeneity of surface characteristics not only weaken the spatial representativeness of point-scale measurements across adjacent areas, but also pose significant maintenance challenges for long-term stable measurements~\citep{collow2012evaluation,loew2008impact,njoku2003soil,peng2017review}. Reanalysis datasets offer excellent spatio-temporal continuity and spatial coverage, but have received less validation against in-situ measurements compared to remote sensing data. Given these limitations, microwave remote sensing techniques have become a reliable and promising approach for global SM retrieval due to their lower uncertainty and reduced bias compared to reanalysis products~\citep{gao2022deep,petropoulos2015surface,zeng2015evaluation}.

Remote sensing (RS) retrieves SM by analyzing electromagnetic radiation reflected or emitted from the land surface, utilizing physical or empirical relationships between sensor measurements and surface parameters~\citep {gao2022deep,zeng2015evaluation}. Among various RS techniques, microwave sensors can penetrate through clouds and operate under all weather conditions, enabling continuous monitoring of SM within the top 5 cm of soil. Compared to optical sensors, passive microwave systems are less affected by surface roughness variations and offer more reliable temporal consistency for long-term observations. Several dedicated SM satellite missions have been deployed, including the advanced microwave scanning radiometer for earth observing system (AMSR-E), its successor AMSR2, the soil moisture and ocean salinity (SMOS) and soil moisture active passive (SMAP), These missions operate at different microwave frequencies and spatial resolutions, employing varying observation geometries to retrieve SM. However, microwave-derived SM products suffer from coarse spatial resolution (typically tens of kilometers) and multi-day revisit cycles\citep{abowarda2021generating}, which constrain their applicability in domains requiring fine-scale data such as carbon cycle research, drought and flood monitoring, water resource management, precision agriculture, extreme climate monitoring, and ecosystem dynamics studies~\citep{carbone1985doppler,fang2018amsr2,li2021high,nagy2024precision,peng2021roadmap,piles2016towards}.

To fill this gap, researchers have developed various approaches to downscale SM products. Current downscaling methods are broadly classified into three types: model-based approaches, active and passive microwave data fusion approaches, and optical/thermal and microwave data fusion approaches. Model-based approaches utilize statistical modeling and land surface process simulation to downscale SM data~\citep{drusch2005observation,ines2002inverse,kaheil2008downscaling,kim2002downscaling,kroes2000integrated,mascaro2011soil,peng2013representative,shin2013development,verhoest2015copula}. Active and passive microwave data fusion methods combine passive sensors' global observations with active sensors' fine spatial resolution. However, these methods are constrained by infrequent radar revisits and geometric registration challenges~\citep{aiazzi2002context,montzka2016investigation,njoku2002observations,wagner2008temporal,wigneron2003retrieving,xing2025retrieval,zhan2006method}. The third category, optical/thermal and microwave data fusion methods, enhance the spatial resolution of coarse microwave SM products by leveraging high-resolution optical and thermal observations. Empirical polynomial fitting downscaling methods based on the universal triangle theory have been developed~\citep{carlson1994method,chauhan2003spaceborne}. Subsequent research has focused on developing soil and vegetation indices, exploring combinations of auxiliary factors, and optimizing algorithms~\citep{merlin2010improved,nasta2018downscaling,piles2011downscaling,zhao2013downscaling}. Additionally, researchers have developed methods based on evapotranspiration process decomposition~\citep{malbeteau2016dispatch,merlin2012disaggregation,molero2016smos}, techniques for establishing soil indices and SM correlation patterns~\citep{jiang2003intercomparison,kim2011improving,peng2015spatial}, and downscaling approaches utilizing artificial intelligence (AI) technology~\citep{srivastava2013machine}. 

In recent years, AI-based methods have been widely applied to the downscaling and reconstruction of SM due to their capacity to capture complex spatio-temporal nonlinear dependencies between auxiliary variables and SM. \cite{srivastava2013machine} pioneered the application of machine learning techniques to SM downscaling, utilizing data from the moderate resolution imaging spectroradiometer (MODIS).

\begin{itemize}

\item Within the domain of machine learning, random forest (RF) and support vector machines (SVMs) have been widely applied in SM downscaling and meteorological variable reconstruction~\citep{abbaszadeh2019downscaling,hutengs2016downscaling,liu2017comparison,im2016downscaling}. Other machine learning techniques such as gradient boosting decision trees (GBDT) and eXtreme gradient boosting (XGBoost) have also shown competitive performance in pixel-level SM downscaling~\citep{abowarda2021generating,im2016downscaling,lei2022quasi,liu2020generating,liu2017comparison,wei2019downscaling,zhao2018spatial,he2025smpd}. These methods effectively capture complex nonlinear relationships between input features and target variables; however, they operate on individual pixels without considering spatial dependencies or temporal continuity. Since SM exhibits strong spatial autocorrelation and temporal persistence, these pixel-wise approaches frequently yield spatially discontinuous and temporally inconsistent results. Methods that incorporate spatio-temporal context are therefore essential for generating coherent predictions~\citep{liu2018understanding, luo2016understanding}.

\item Within the domain of deep learning, methods such as artificial neural networks (ANN), long short-term memory networks (LSTM), and convolutional neural networks (CNN) have demonstrated advantages in various aspects such as modeling complex nonlinear relationships, capturing temporal dependencies, and extracting spatial features in SM downscaling~\citep{fang2017prolongation,orth2021global,rodriguez2015soil,sobayo2018integration}. These deep learning approaches demonstrate significant potential for advancing downscaling methodologies. For example,~\cite{kolassa2018estimating} constructed an ANN to map global surface SM from SMAP observations, demonstrating strong data assimilation capabilities, and~\cite{gao2022deep} collected data from the international soil moisture network (ISMN) to train a deep neural network (DNN), achieving performance that surpasses SMAP products.

\end{itemize}

The Tibetan Plateau (TP)~\citep{qiu2008china} is a highly sensitive region to climate change~\citep{yang2013multiscale}, playing a pivotal role in shaping the climate evolution of the Asian monsoon and atmospheric circulation processes in the Northern Hemisphere through complex interactions among land, atmosphere, and ocean~\citep{bothe2011large,jin2005impacts,xie2023oceanic}. However, the widespread distribution of permanent and seasonal permafrost across the TP~\citep{abowarda2021generating}, coupled with harsh natural conditions, results in sparse ground observation stations that are difficult to maintain, leading to severe deficiencies in in-situ SM measurements for this region. Consequently, there is an urgent need to develop high-precision SM downscaling technologies. Therefore, \cite{jiang2017evaluation} employed a back propagation neural network (BPNN) algorithm with five microwave SM products, demonstrating that BPNN outperforms traditional regression techniques and yields superior correlation coefficients (R) with in-situ SM data over the TP. Moreover, \cite{wei2019downscaling} and \cite{zhao2022downscaling} utilized the GBDT method in conjunction with 26 indices related to varying levels of vegetation water content or SM, and compared its performance with deep belief networks (DBN), ResNet, BPNN, and RF, successfully downscaling SMAP data on the TP from a 36 km to 1 km resolution. Alternatively, \cite{shangguan2023inter} introduced the MATCH method, which enhances the downscaling performance of ANN, RF, ResNet, and LSTM through a combined machine learning approach for TP applications. More recently, \cite{shangguan2024long} constructed five machine learning downscaling models annually and minimized the error variance of SM data to merge the datasets, resulting in daily 1 km resolution SM data on the TP.

Constrained by the inadequate temporal resolution of satellite-derived surface auxiliary variables, most existing studies have focused on the spatial downscaling of SM. Reanalysis datasets generate hourly-resolution surface and near-surface parameters through data assimilation and numerical modeling. By utilizing validated surface auxiliary variables from reanalysis data to compensate for the temporal resolution limitations of satellite-derived surface auxiliary variables, this strategy provides the data foundation for multi-scale datasets for spatio-temporal downscaling. However, the integration of multi-source data with heterogeneous spatio-temporal characteristics necessitates specialized network architectures. To address this challenge, we first propose the progressive scale convolutional network (PSCNet) that systematically handles the trade-off between spatial scales and feature dimensions. We subsequently introduce a temporal convolutional module to enhance the network's temporal representation capability. Finally, through validation with original microwave satellite data, in-situ measurements, temporal generalization testing, and temporal dynamics visualization, our method demonstrates competitive results compared to existing approaches. The contributions of our research can be summarized as follows:

\begin{itemize}
  \item We pioneered the integration of validated high spatio-temporal resolution reanalysis variables for spatio-temporal downscaling on the TP, achieving results at 10-km and 3-hour resolution. Furthermore, we demonstrated the feasibility and effectiveness of utilizing multi-temporal auxiliary variables within this framework.
  
  \item We conducted a systematic comparative analysis of spatially explicit models versus point-based models for SM reconstruction through multi-perspective validation including in-situ site validation, satellite data validation, inter-model validation, and visual assessment.
  
  \item We developed a separable spatio-temporal convolutional framework and introduced PSCNet, specifically designed for downscaling tasks with end-to-end spatio-temporal modeling capability that supports arbitrary input sizes, ensuring spatial coherence and eliminating tile boundary artifacts.
  
\end{itemize}

\section{Study Area and Data}
\subsection{Study Area}
The TP is the highest and largest plateau in the world, covering an area of about $3.08 \times 10^6$ square kilometers, with an average elevation of about 4320 m. It is situated between latitudes $26^\circ$ and $40^\circ$ north and longitudes $67^\circ$ and $104^\circ$ east. TP lies at the convergence of Asian mountain ranges, bounded by the Kunlun-Qilian Mountains in the north, the Himalayas in the south, the Karakoram Range and the Pamir Plateau in the west, and the Hengduan Mountains in the east, serving as the birthplace of many major Asian rivers~\citep{zhang2021redetermine}. It is also a region with complex climatic conditions and an uneven distribution of vegetation. It's permafrost and seasonally frozen ground cover more than 40\% and 56\%, respectively~\citep{zou2017new}, while SM is highly heterogeneous due to its complex topography and diverse land cover.

\begin{figure}
	\centering
	\includegraphics[width=.9\textwidth]{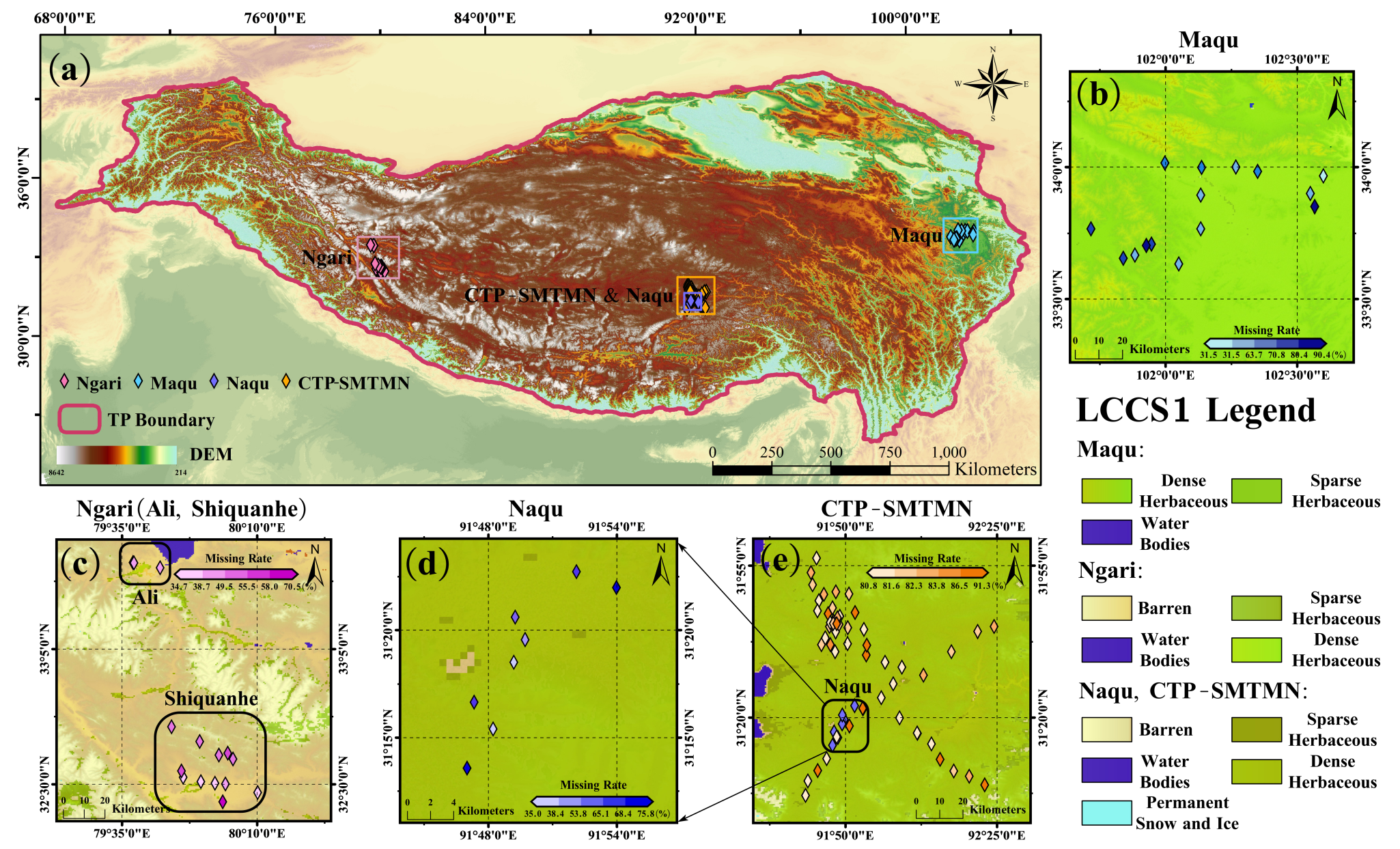}
	\caption{Elevation distribution and observation networks over the TP. (a) Overview showing elevation and four networks. (b-e) Site distribution with land cover types for Maqu, Ngari, Naqu, and CTP-SMTMN (abbreviated as CTP) networks, respectively. Symbol colors denote data missing rates. Background combines land cover types (30\% transparency) with DEM topography. Legend describes FAO land cover classification system 1 (LCCS1) types within each network.}
	\label{FIG:1}
\end{figure}

\subsection{Data}
Surface radiation, meteorological factors, topography, vegetation, and soil characteristics are considered important SM inversion~\citep{crow2012upscaling,peng2016comparison,zhao2013downscaling}. Therefore, we synthesized multiple satellite products, including SMAP L3 SM product (SPL3SMP, Version 8), MODIS land surface parameters, and topographic data from Hydrological data and maps based on shuttle elevation derivatives at multiple scales (HydroSHEDS), combined with climate data from european centre for medium-range weather forecasts (ECMWF) distributing ERA5-Land hourly reanalysis products and gauge-adjusted precipitation data provided by the global satellite mapping of precipitation (GSMaP). Table~\ref{tbl1} details the specific datasets used to construct the multi-temporal scale training dataset.

\begin{table}[t]
  \caption{List of surface variables, data sources, and spatial and temporal resolutions used to construct the multi-source dataset.}
  \label{tbl1}
  \begin{tabular*}{\textwidth}{l|l|l|l|l}
    \toprule
    \textbf{Type} & \textbf{Variable name} & \textbf{Source} & \textbf{Temporal resolution} & \textbf{Spatial resolution} \\
    \midrule
    \textbf{Interpolation variable} & SM & SPL3SMP & Descending/Ascending Pass & 36km \\
    \midrule
    \multirow{12}{*}{\textbf{Auxiliary Dynamic Variables}} & HOY & — & Derived & — \\
    & LST-Day & MYD11A1 & Daily & 1km \\
    & LST-Night & MYD11A1 & Daily & 1km \\
    & Albedo-BSA & MCD43C3 & Daily & 5km \\
    & Albedo-WSA & MCD43C3 & Daily & 5km \\
    & NDVI & MYD13A1 & 16 Day & 500m \\
    & EVI & MYD13A1 & 16 Day & 500m \\
    & Precipitation & GSMaP-Gauge & Hourly & 10km \\
    & Flux & ERA5-Land & Hourly & 10km \\
    & Snowmelt & ERA5-Land & Hourly & 10km \\
    & Runoff & ERA5-Land & Hourly & 10km \\
    & Surface radiation & ERA5-Land & Hourly & 10km \\
    & Evaporation & ERA5-Land & Hourly & 10km \\
    \midrule
    \multirow{8}{*}{\textbf{Auxiliary Static Variables}} & Longitude & — & Static & — \\
    & Latitude & — & Static & — \\
    & Clay & SoilGrids250m 2.0 & Static & 250m \\
    & Sand & SoilGrids250m 2.0 & Static & 250m \\
    & Silt & SoilGrids250m 2.0 & Static & 250m \\
    & SOC & SoilGrids250m 2.0 & Static & 250m \\
    & DEM & HydroSHEDS & Static & 30m \\
    & Slope & HydroSHEDS & Static & 30m \\
    & Aspect & HydroSHEDS & Static & 30m \\
    \midrule
    \multirow{2}{*}{\textbf{Land Cover}} & LC-Prop1:LCCS1 & MCD12Q1 & Yearly & 500m \\
    & K-GCC & — & Static & 1km \\
    \bottomrule
  \end{tabular*}
\begin{tablenotes}
    \footnotesize
    \item[*] HOY: Hour of Year, LST: Land Surface Temperature, NDVI: Normalized Difference Vegetation Index, EVI: Enhanced Vegetation Index, K-GCC: Köppen-Geiger Climate Classification, DEM: Digital Elevation Model, SOC: Soil Organic Carbon Content.
\end{tablenotes}
\end{table}

\subsubsection{SMAP Soil Moisture Data}
The SMAP satellite, launched by NASA on January 31, 2015, was designed to retrieve global surface SM and freeze/thaw state. It carries both an active L-band microwave radar and a passive radiometer. After the radar failed, the radiometer continues operations in a sun-synchronous orbit with equatorial crossings at 6:00 AM (descending) and 6:00 PM (ascending) local solar time, achieving global coverage every 2-3 days. The vast permanent and seasonal permafrost region on the TP presents challenges for L-band radiometer SM retrieval, resulting in extensive gaps in SPL3SMP across this region (Fig.~\ref{FIG:1}). This study utilizes the SMAP L3 radiometer-derived product SPL3SMP (Version 8), which provides twice-daily global soil moisture data at a 36 km resolution on an EASE-Grid. During processing, we filtered data to remove observations flagged for frozen conditions (surface temperature less than 273.15 K), open water coverage >10\%~\citep{lei2022quasi}, and values failing the valid minimum threshold (< 0.02). Data were downloaded following reprojection to a geographic coordinate system using NASA Earthdata Search's reprojection tool.

\subsubsection{MODIS Auxiliary Data}
MODIS, aboard the Terra and Aqua satellites, is widely used for global environmental monitoring~\citep{ustin2021current}. This multispectral sensor covers the visible, near-infrared, and thermal infrared spectral bands. Its data for Collection 6 have completed the third phase of quality validation~\citep{skakun2018transitioning}. The surface variables employed during the experiment comprised LST, albedos, NDVI, and EVI, supplemented by the auxiliary LCCS1 variable for land cover classification analysis.

LST is derived from the global 1 km resolution MYD11A1 product from the Aqua satellite, utilizing both daytime and nighttime surface temperature bands. Albedos are calculated using the daily MCD43C3 product, which integrates 16 days of Terra and Aqua MODIS data and employs all bands from both white-sky and black-sky observations. NDVI and EVI, on the other hand, are derived from MYD13A1 data synthesized using the 16-day maximum value compositing (MVC) method at a 500 m spatial resolution. MCD12Q1 provides annual land cover types at a 500-meter resolution from 2001 to 2022. The data acquisition period for all of the aforementioned datasets spans from January 1, 2016, to January 1, 2019. All of these datasets are freely accessible on the Google Earth Engine.

\subsubsection{ERA5-Land Auxiliary Data}
The ERA5-Land dataset, released by ECMWF, represents advancements in surface physics modeling and multi-source data assimilation techniques, providing hourly data from 1950 to the present at a spatial resolution of $0.1^\circ \times 0.1^\circ$ with global land surface coverage.
ERA5-Land is produced by running the ECMWF land surface model (H-TESSEL) forced by downscaled meteorological fields from the ERA5 climate reanalysis, with elevation corrections for thermodynamic variables. The accuracy of ERA5-Land surface variables has been extensively validated across different regions~\citep{crowhurst2021contrasting,gualtieri2021reliability,hersbach2016era5,hua2019assessing,zhang2022dataset}. For SM downscaling in this study, we selected ERA5-Land surface variables specifically validated over the TP region and frequently used in SM retrieval. These variables include surface temperature (1-7 cm), surface solar radiation, surface net radiant heat flux, surface latent heat flux, surface sensible heat flux, and total evaporation.~\citep{dong2020comparison,li2023evaluation,martens2020evaluating,yilmaz2023accuracy,zhang2020evaluation,zheng2022assessment}.

\subsubsection{Other Auxiliary Data}
For comprehensive SM modeling, we incorporated additional precipitation and soil property datasets. Precipitation is a primary controlling factor for the abrupt changes in SM during short periods. High-temporal-resolution precipitation data enable accurate characterization of rapid SM variations~\citep{dai2022soil,wang2018spatiotemporal,yin2023spatiotemporal}. The GSMaP-Gauge Precipitation dataset~\citep{jiang2023evaluation} provides global data at $0.1^\circ$ resolution at half-hourly intervals. Compared to other products in the GPM series, GSMaP-Gauge exhibits the highest accuracy in mainland China~\citep{kubota2020global,zhou2020comprehensive}. Soil texture (clay, sand, silt, SOC) provides a fundamental description of the soil's hydraulic conductivity and permeability~\citep{nemes2005influence}. Combined with topographic factors (DEM, slope, and aspect), these variables can serve as static features to construct a soil hydrodynamic model. Accordingly, we selected the global SoilGrids250m soil property dataset and the HydroSHEDS DEM product. SoilGrids250m offers global predictions of soil properties and classifications at a 250m resolution, while HydroSHEDS 3 arc-second enhances the global shuttle radar topography mission (SRTM)  product with hydrological feature optimization and custom corrections such as void-filling, filtering, stream burning, and upscaling techniques~\citep{lehner2008new}, which effectively address the notable radar shadow effects in mountainous and aquatic zones. We employ the HydroSHEDS 3 arc-second Void-filling DEM product and derive slope and aspect values through terrain analysis.

\subsubsection{In-situ Measurements}\label{sec:in_situ_measurements}
The ISMN is a centralized global SM data system maintained by the Vienna University of Technology~\citep{dorigo2017esa}. This system provides hourly site-based observation data in UTC and has been widely validated~\citep{ochsner2013state}. Our precision assessment relies on site data from five in-situ measurement networks on the TP: Naqu, Maqu, two Ngari sub-networks (Ali and Shiquanhe, treated separately due to their significant spatial separation, shown in Fig.~\ref{FIG:1}), and CTP, as detailed in Table~\ref{tbl2}.

\begin{table}[ht]
  \begin{threeparttable} 
  \caption{Summary of in-situ observation networks used for validation.}
  \label{tbl2}
  \begin{tabular*}{\textwidth}{lcccc}
  \toprule
  \textbf{Network} & \textbf{Number of stations} & \textbf{\makecell{ Climate \\ classification*}} & \textbf{\makecell{Data period \\ (valid: Apr-Oct)}} & \textbf{\makecell{Missing data rate \\ (3-hour intervals)}} \\
  \midrule
  Naqu & 9 & \makecell{Dwd: Cold, dry winter, \\ very cold winter} & \makecell{2016.01.01- \\ 2018.10.31} & 59.6\% \\
  CTP & 52 & ET: Polar, tundra & \makecell{2016.01.01- \\ 2016.09.20} & 83.6\% \\
  Ali & 3 & BWk: Arid, desert, cold & \makecell{2016.01.01- \\ 2018.10.31} & 46.5\% \\
  Shiquanhe & 12 & BWk: Arid, desert, cold & \makecell{2016.01.01- \\ 2018.10.31} & 47.5\% \\
  Maqu & 15 & ET and Dwd & \makecell{2016.01.01- \\ 2018.10.30} & 66.9\% \\
  \bottomrule
  \end{tabular*}
  \begin{tablenotes}
    \footnotesize
    \item[*]Climate classification based on Köppen-Geiger Global 1 km climate classification maps~\citep{beck2023high}.
\end{tablenotes}
\end{threeparttable}
\end{table}

Systematic differences in measurement results from satellites and sparse observation networks are influenced by factors associated with different measurement modes, such as spatial scales, observational mechanisms, probing depths, and sensitivity to instantaneous rainfall~\citep{colliander2017validation,parinussa2011error}. Furthermore, these networks are located on seasonal permafrost~\citep{zou2017new}, presenting significant challenges to data integrity and reliability. Therefore, to obtain robust site observation results, the following data processing rules have been established. 1) Select sites with the shallowest surface depth of less than 5cm. 2) Use ISMN data marked as "Good"~\citep{dorigo2013global}. 3) Only select data from the melt season between April 1 and November 1 from 2016 to 2019 for verification, excluding sites where the data missing rate exceeds 95\%. 4) When more than two sites are located within the same grid point, select the most representative site to reduce the disparity of SM assessment weights~\citep{dorigo2015evaluation,zhang2019comprehensive}.

\subsection{Data Processing}
In this study, we integrate multi-source datasets with heterogeneous spatio-temporal resolutions to construct a temporally unified dataset that spans from 03:00 on January 1, 2016, to 21:00 on December 31, 2018. To meet the requirements of spatial downscaling, we further organize the data into two distinct datasets: a 36km training dataset (w/ SM) and a 10km inference dataset (w/o SM), both at 3-hour temporal intervals

For the general data processing workflow, initial processing of different data types involves cropping, stitching, quality control, and reprojection to unify coordinate systems to the WGS84 system. Subsequently, spatial resolutions are standardized using bilinear interpolation to resample the data onto regular latitude-longitude grids at resolutions of 36km and 10km to ensure alignment with the resolutions of SMAP SM data and ERA5-Land data, respectively. Finally, all auxiliary datasets were temporally resampled to a 3-hour resolution using interpolation methods. The SMAP SM data, serving as the training target, was kept at its native observation times of approximately 06:00 and 18:00 local solar time daily. To establish a uniform temporal framework for the analysis, these local solar time observations are subsequently referenced by the nominal 06:00 and 18:00 UTC timestamps. Additionally, we incorporate longitude and latitude as spatial context variables and HOY as a temporal context variable to enhance model performance.

For model training, we first apply Z-Score normalization to the 36km training dataset, then standardize the 10km inference dataset using the same statistical parameters. Training samples are extracted as 32x32 spatial patches with a temporal length of 5 and a stride of 10 pixels. The training data is partitioned into training (70\%), validation (15\%), and test (15\%) sets. Data augmentation techniques, including flipping, transposing, and mirroring, are applied to the training samples.

For ground truth validation, ISMN site data is preprocessed and temporally aggregated to a 3-hour resolution.

\section{Methodology}
\begin{figure}
	\centering
	\includegraphics[width=.9\textwidth]{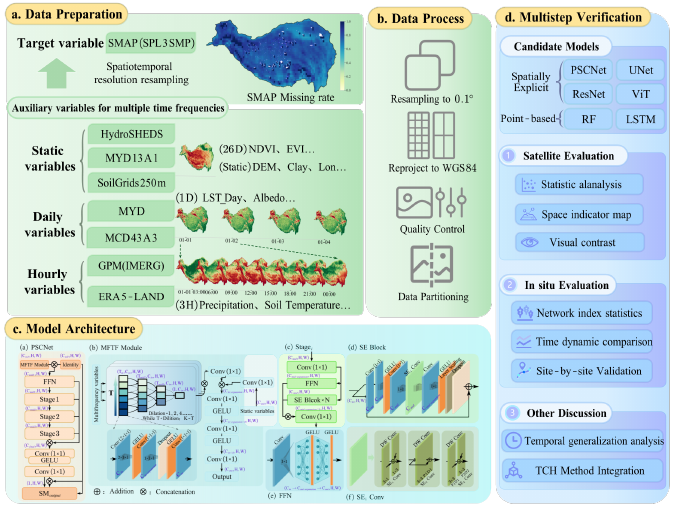}
	\caption{Overall framework and methodological workflow of SM spatial downscaling.}
	\label{FIG:2add}
\end{figure}

The downscaling framework and experimental procedure in this study are illustrated in Fig.~\ref{FIG:2add}, comprising four main modules: (a) Data preparation, which integrates multi-source datasets with heterogeneous spatio-temporal resolutions. (b) Data processing involves reprojecting, resampling, quality control, and data clipping steps. (c) The proposed PSCNet consists of the multi-frequency temporal fusion (MFTF) module and the squeeze-and-excitation (SE) block. (d) A multi-step validation of the downscaled products, where the models are divided into two categories: spatially explicit models (e.g., PSCNet, UNet) and point-based models (e.g., LSTM, RF). The comprehensive quality assessment is carried out through various perspectives, including in-situ site validation, satellite data validation, and inter-model validation, among others.

\subsection{Problem Formulation}
The SM downscaling process can be modeled as the learning and transfer of spatio-temporal mapping relationships using multi-source auxiliary variables. Given the original multi-source auxiliary dataset $\mathcal{D} = \{\mathcal{D}_1, \mathcal{D}_2, \dots, \mathcal{D}_n\}$, where each $\mathcal{D}_k$ has unique spatio-temporal resolution, spatio-temporal alignment generates a unified low-resolution dataset $\mathcal{D}_L = \{X_L, Y_L\}$ (containing auxiliary variables $X_L$ and SM ground truth $Y_L$) and a high-resolution dataset $\mathcal{D}_H = \{X_H\}$ (containing only auxiliary variables $X_H$). The extracted spatio-temporal data block $x^{(i)} \in \mathbb{R}^{T \times H \times W \times C}$ characterizes region $i$'s multi-variable state over a temporal sequence of length $T$, where $(H, W)$ is the spatial scale and $C$ is the variable dimension. In the low-resolution domain, we construct a network model $f_\theta$ to learn relationships between SM and auxiliary variables, where the parameter set $\theta = \{\theta_{\text{temp}}, \theta_{\text{spat}}\}$ models the spatio-temporal dimensions: The temporal encoder $f_{\theta_{\text{temp}}}$, implemented as our MFTF module, aggregates the time series into contextual features, then the spatial decoder $f_{\theta_{\text{spat}}}$ reconstructs these features into SM predictions:
\begin{eqnarray}
\hat{Y}_L^{(i)} = f_{\theta_{\text{spat}}} \left( f_{\theta_{\text{temp}}} \left( \{x_{L,1}^{(i)}, \dots, x_{L,T}^{(i)}\} \right) \right) \in \mathbb{R}^{H_L \times W_L}
\end{eqnarray}
The model is optimized by minimizing the loss function:
\begin{eqnarray}
\theta^* = \underset{\theta}{\operatorname{arg\,min}} \sum_i \mathcal{L} \left( \hat{Y}_L^{(i)}, Y_L^{(i)} \right)
\end{eqnarray}
Finally, the trained model enables cross-scale transfer by applying the learned spatio-temporal mapping $f_{\theta^*}$ to the high-resolution auxiliary variables $X_H$ to generate high-resolution SM:
\begin{eqnarray}
\hat{Y}_H^{(i)} = f_{\theta^*} \left( X_H^{(i)} \right) \in \mathbb{R}^{H_H \times W_H}
\end{eqnarray}

\subsection{Model Architecture}
\begin{figure}
	\centering
	\includegraphics[width=1.0\textwidth]{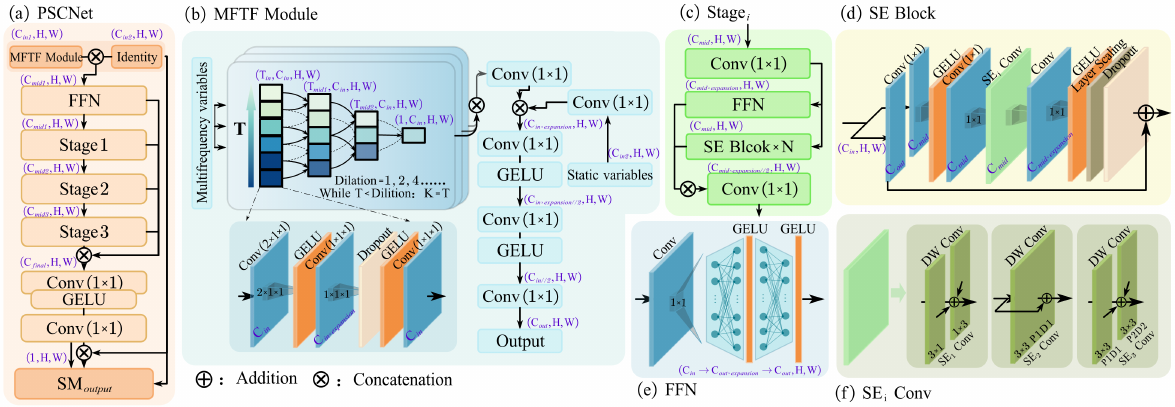}
	\caption{The architecture of the proposed PSCNet. The main figure (a) illustrates the overall structure, while (b-f) provide detailed views of its key components: the MFTF Module, the Stage block, the SE block, the feed-forward network (FFN), and the SE Convolutional layer.}
	\label{FIG:3}
\end{figure}

Achieving effective spatio-temporal mapping in practice requires careful consideration of network architecture design. The key challenge in downscaling network design is to preserve cross-scale mapping relationships while flexibly capturing inter-variable dependencies across dimensions. Moreover, robust spatio-temporal modeling is essential for ensuring the spatio-temporal consistency of the results, which in turn hinges on efficient information distillation and redundancy removal. To address these challenges, we examine the network architecture from several perspectives:

\begin{itemize}

\item Maintaining Spatial Integrity without Downsampling. The U-shaped architecture is widely used in modern CNN designs, where progressive downsampling enhances the model's ability to capture spatial hierarchies. However, even with skip connections, downsampling operations inevitably lead to the loss of fine-grained spatial details essential for high-resolution reconstruction. Additionally, patch-wise reconstruction can introduce boundary artifacts and spatial discontinuities. To address these issues, we adopt a convolutional network architecture that omits downsampling operations, thereby preserving the completeness of spatial information.

\item Stabilizing Generalization without Batch Normalization. In the downscaling process, data sampling demonstrates strong sequential dependency. Batch normalization (BN) may introduce variance shifts that result in numerical instability during inference~\citep{li2019understanding}. To mitigate this issue, we remove BN layers from the network to enhance stability.

\item Enhancing Information Retention with Gaussian Error Linear Unit (GELU) and Inverted Residuals. Activation functions play a crucial role in enabling ANNs to capture nonlinear relationships~\citep{glorot2010understanding}. ReLU activations can cause information loss through hard thresholding, particularly problematic when preserving subtle spatial patterns in downscaling tasks~\citep{howard2017mobilenets}. To address this limitation, we replace all ReLU layers with GELU activations and incorporate inverted residual structures to maintain richer feature representations~\citep{liu2022convnet}.

\item Reconstructing Inference Patterns with Hybrid Multi-Level Network Architectures. Localized feature extraction demonstrates strong performance in downscaling tasks~\citep{abbaszadeh2019downscaling,kim2018intercomparison,zhuo2017multi}; however, spatio-temporal modeling with larger receptive fields yields superior spatial coherence. To combine the strengths of both approaches, we design a multi-stage hybrid architecture embodied in our Stage module (Fig.~\ref{FIG:3} (c)), which explicitly couples a FFN for localized inference with a convolutional branch built upon SE Blocks to extract hierarchical spatial features~\citep{ding2024unireplknet}. Dilated and factorized convolutions further enlarge the receptive field efficiently while preserving model compactness.

\item Modeling spatio-temporal Dependencies.
Effective spatio-temporal downscaling hinges on the model's ability to capture complex, non-linear dependencies. Our MFTF module (Fig.~\ref{FIG:3} (b)) achieves this by first establishing an absolute spatio-temporal context using positional encodings (Hour of Year, longitude, and latitude). This context-aware data is then processed by a temporal convolutional network (TCN)~\citep{bai2018empirical}, which serves as the core engine for dependency modeling. Its capacity is enhanced by internal feature enrichment layers, while no-padding convolutions are used to progressively distill the learned temporal relationships across various scales into a final, feature-rich spatial representation. If the length of the current time block, $T_l$, is smaller than the dilation length, $D_l$, a convolution kernel of length $T_l$ is used to reduce the temporal dimension to 1.

\end{itemize}
Collectively, these design principles culminate in the PSCNet architecture, illustrated in Fig.~\ref{FIG:3}.

\subsection{Loss Function}
Patch-based training enhances spatial coherence within patches but frequently introduces artifacts along patch boundaries. Therefore, we implement a multi-task loss function combining RMSE and structural similarity index measure (SSIM) to produce downscaled SM data better aligned with human visual perception.

RMSE quantifies pixel-level errors. To prioritize edge regions and mitigate mosaic artifacts during reconstruction, an edge-weighting kernel $W_e$ is incorporated into the $L_{\mathrm{RMSE}}$ calculation. The kernel $W_e(i,j)$, and $L_{\mathrm{RMSE}}$ are defined as follows:
\begin{eqnarray}
  W_e\left( i,j \right) =1+\left( \mathrm{ratio}-1 \right) \cdot \frac{2\bullet \sqrt{\left( i-\frac{H-1}{2} \right) ^2+\left( j-\frac{W-1}{2} \right) ^2}}{\sqrt{H^2+W^2}}
\end{eqnarray}

\begin{eqnarray}
  L_{\mathrm{RMSE}}=\sqrt{\frac{1}{H\times W}\sum_{i=1}^H{\mathrm{}}\sum_{j=1}^W{\mathrm{}}W_e\left( i,j \right) \cdot \left( \hat{y}\left( i,j \right) -\mathrm{y}\left( i,j \right) \right) ^2}
\end{eqnarray}
Here, $\hat{y}$ and $y$ denote the predicted and true SM values, while $H$ and $W$ specify the height and width of the patch. The parameter ratio is the weighting coefficient.

SSIM assesses perceptual quality based on human vision and exhibits strong generalizability~\citep{wang2004image}. Incorporating SSIM into a hybrid loss function effectively suppresses reconstruction artifacts and improves visual quality~\citep{huang2020deep}. The $L_{\mathrm{SSIM}}$ and composite loss $L_{\mathrm{FULL}}$ are defined as:
\begin{eqnarray}
L_{\mathrm{SSIM}}=1-\frac{\left( 2\mu _{\hat{y}}\mu _y+C_1 \right) \left( 2\sigma _{\hat{y}y}+\mathrm{C}_2 \right)}{\left( {\mu _{\hat{y}}}^2+{\mu _y}^2+C_1 \right) \left( {\sigma _{\hat{y}}}^2+{\sigma _y}^2+C_2 \right)}
\end{eqnarray}
\begin{eqnarray}
L_{\mathrm{FULL}}=\alpha L_{\mathrm{RMSE}}+\left( 1-\alpha \right) L_{\mathrm{SSIM}}
\end{eqnarray}
where $\mu_{\hat{y}}$ and $\mu_{y}$ are the means of the predicted and true SM values, respectively, while $\sigma_{\hat{y}}^2$ and $\sigma_{y}^2$ are their variances, $\sigma_{\hat{y}y}$ is the covariance, and $C_1$ and $C_2$ are small stabilizing constants. The hyperparameter $\alpha$ controls the blending ratio. Statistics ($\mu, \sigma^2, \sigma_{xy}$) are computed over the model's output patches.

\section{Experimental Results}
The following sections outline the evaluation scenarios: 1) a downscaling quality assessment of SM inversion products based on SMAP microwave satellite data, 2) an assessment of the feasibility of reconstructing high-temporal resolution data using in-situ measurements, and 3) a visual assessment of downscaled products.

This study employs four commonly used evaluation metrics for SM downscaling~\citep{liu2020generating,qu2021inter}, namely the R, bias, RMSE, and ubRMSE. The specific formulas for these metrics are provided in \ref{sec:statistical_indicators}.

\subsection{Model Implementation and Configuration}
The models used in this study are primarily divided into two categories: 1) spatially explicit models that leverage contextual spatial information, including models designed to capture spatial dependencies, such as ResNet, UNet, and Vision Transformer, as well as our proposed spatio-temporal model, PSCNet; and 2) point-based models, which treat each location independently, including temporal sequence regression models such as LSTM and pixel-wise methods like RF. 

For the spatially explicit models, during training, we apply progressive masking to SM data, transitioning from dual-task learning (variable inference and missing data reconstruction) to pure variable inference to enhance spatial reasoning capabilities~\citep{papamakarios2017masked,perez2017effectiveness}. Furthermore, during inference, the PSCNet architecture is capable of processing inputs of arbitrary size, allowing us to employ a full-image inference mode to reinforce spatial coherence, while other models perform inference on 32×32 patches, consistent with their training configuration. For point-based models, LSTM uses a time window of 56 steps (corresponding to one week), with a hidden layer dimension of 512 and an initial learning rate of 0.0005. For RF, the hyperparameters are set to 300 decision trees, with a maximum depth of 10 for each tree and a minimum sample size of 10 and 6 for node splitting and leaf nodes, respectively. Finally, to ensure fair performance comparison in downscaling tasks, we align all models to comparable FLOPS or parameter counts.

\subsection{Model Evaluation Using SMAP Microwave Satellite Data}
To evaluate the downscaled products, we first spatially aggregate them from the 10 km resolution back to the original 36 km SMAP grid. Although spatially aggregating downscaled products back to the original 36 km resolution introduces uncertainty, especially in regions with significant SM variability, this approach remains effective for assessing how well the downscaled products preserve original spatial patterns. During the training phase, the original 36 km SMAP microwave satellite data serve as the low-resolution training target, while in the inference phase, 10 km auxiliary variables are fed into the model to generate 10 km SM products. These products are then aggregated to match the spatial resolution of the original 36 km SMAP data. After addressing missing values, R values are computed at each spatial location. The data are subsequently flattened to calculate statistical metrics between the two datasets, and a scatter density plot is created between the upsampled products and the original SMAP data.

\begin{figure}
	\centering
	\includegraphics[width=1\textwidth]{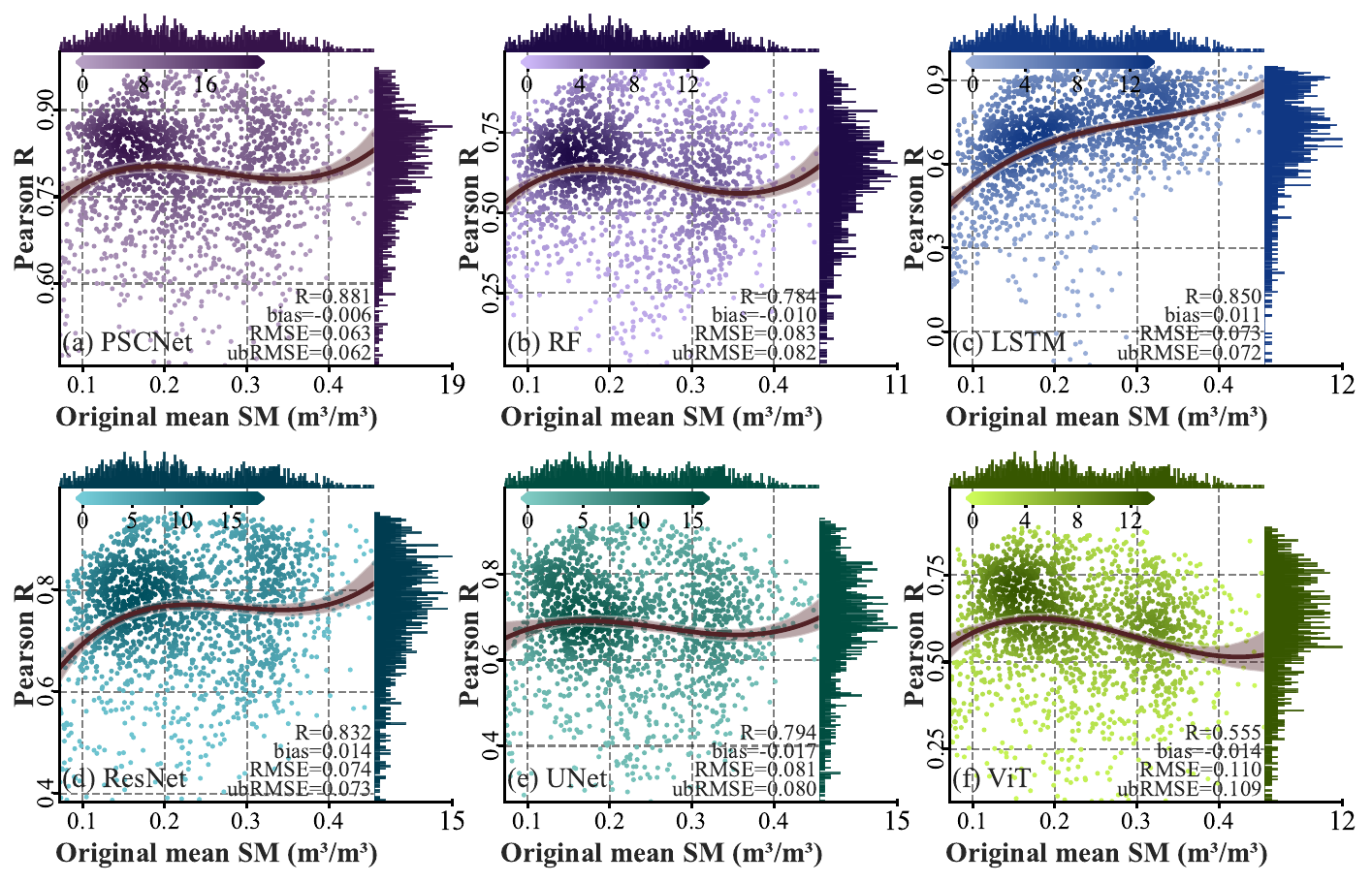}
	\caption{Scatter density plots showing R between downscaled products from ten downscaling methods and original SMAP microwave satellite observations. Overall statistical metrics are displayed in the lower right corner of each subplot. (a) PSCNet; (b) RF; (c) LSTM; (d) ResNet; (e) UNet; (f) ViT.}
	\label{FIG:4}
\end{figure}

Fig.~\ref{FIG:4} shows correlation density scatter plots between downscaled products and original SMAP observations. The mean SM values cluster around 0.2 (m$^3$/m$^3$) and 0.3 (m$^3$/m$^3$), with largely consistent scatter patterns observed across most models in the downscaling task. Among these, the proposed PSCNet model achieves the best performance, with a Pearson R of 0.881, an RMSE of 0.063, and an ubRMSE of 0.062. Among the baseline models, the point-based RF delivers a performance with a Pearson R of 0.784, which is comparable to some models that leverage spatial context, although its overall accuracy is moderate. The LSTM model, in contrast, exhibits a unique performance pattern distinct from all other methods. It shows enhanced accuracy in regions with high SM averages but performs poorly where SM is low, a behavior likely related to its extended training sequence of 56 time steps creating a strong temporal dependency. For the other deep learning baselines designed to capture spatial dependencies, ResNet and UNet achieve the most competitive results with Pearson R values of 0.832 and 0.794 respectively, placing their performance closest to our proposed PSCNet model. Finally, ViT exhibits markedly inferior performance across all key evaluation metrics, exemplified by a low Pearson R of 0.555 and the highest error values among all models.

\begin{figure}
	\centering
	\includegraphics[width=1.0\textwidth]{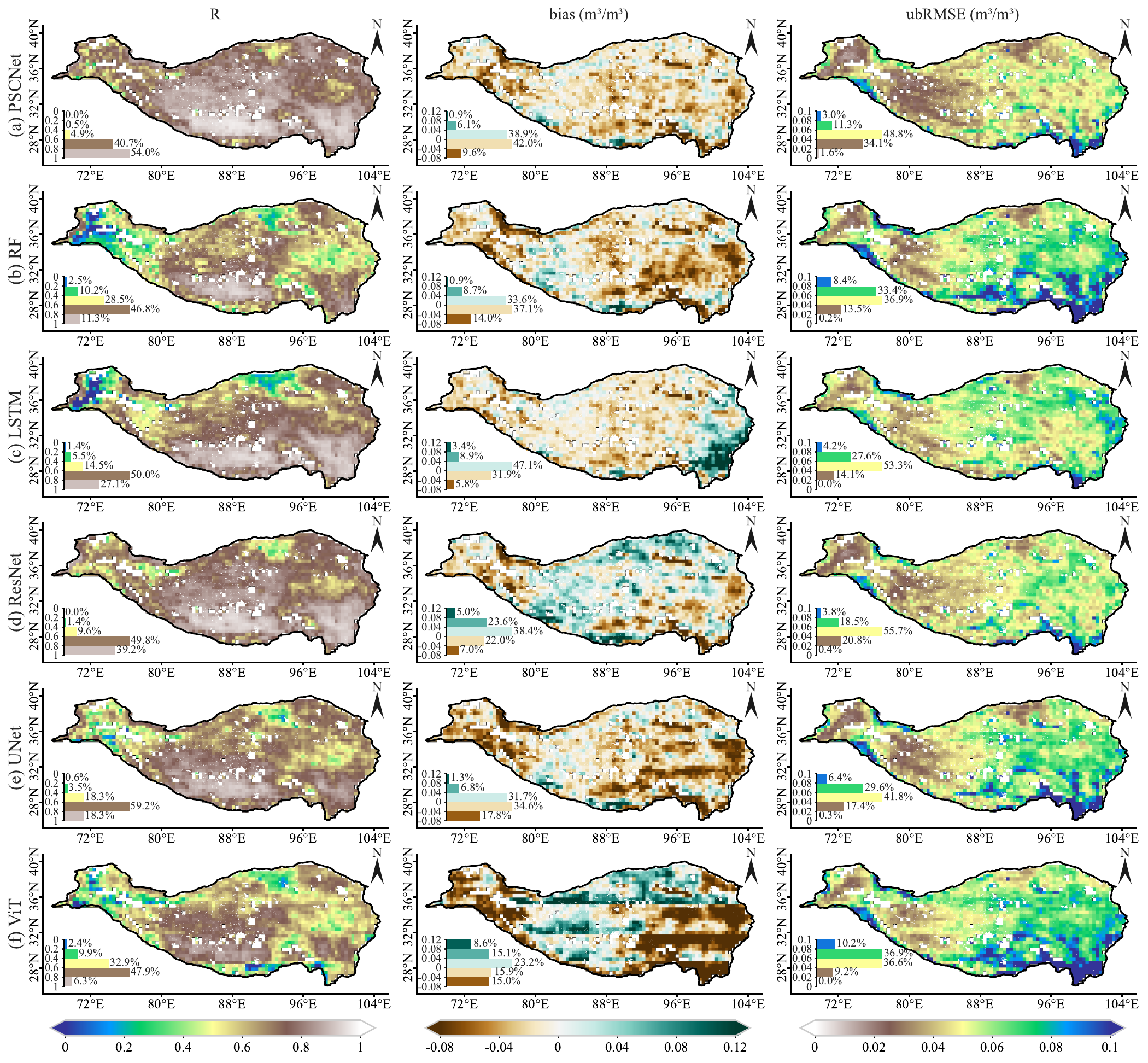}
	\caption{Spatial distribution maps of validation metrics (R, bias, and ubRMSE from left to right) for downscaled products from different models, with histograms showing pixel distribution within each metric range.}
	\label{FIG:7}
\end{figure}

Fig.~\ref{FIG:7} displays the spatial distribution maps of validation metrics for the different downscaled products. The spatial patterns reveal that model performance correlates strongly with regional moisture conditions. This correlation is driven by the region's distinct geography; the Himalayas block moisture-laden South Asian monsoon flows, while the westerlies desiccate leeward regions after crossing the Iranian Plateau. These climatic drivers result in a pronounced southeast-to-northwest SM gradient across the TP, with humid conditions in the southeast giving way to extreme aridity in basins from the Karakoram to the Qaidam. Correspondingly, the spatial patterns of the validation metrics for all models mirror this moisture gradient, consistently showing higher accuracy in the moisture-abundant southeast and degraded performance in the arid northwest.

Specifically, regarding the R metric, all models exhibit weaker performance in the Pamir Plateau and the Qaidam Basin, with the point-based models being the most severely affected, showing significant correlation distortion near the Pamir Plateau. In contrast, PSCNet maintains the highest correlation in these challenging areas. Quantitatively, 94.6\% of the spatial domain for PSCNet has an R value greater than 0.6, with over 54\% of the region surpassing an R of 0.9. Regarding bias, a general overestimation is observed near the Tanggula and Kunlun Mountains, while a general underestimation is prevalent in the northern TP. Among the models, PSCNet and LSTM exhibit the smallest overall bias and show excellent performance. In contrast, RF and UNet generally show an overestimation, while ResNet shows a tendency for underestimation. Furthermore, the results from ViT suffer from severe spatial artifacts, representing the poorest performance. In terms of ubRMSE, all models show higher errors near the Hengduan Mountains. PSCNet achieves the best overall ubRMSE, with only 14.3\% of its domain showing a value greater than 0.06. This is a considerable improvement over the next-best model, ResNet (22.3\%), and is 17.9 percentage points lower than the average of all models (32.2\%).

Overall, among all downscaled products, PSCNet demonstrates the best capability in preserving the features of the original microwave satellite data, achieving the top performance across all metrics in this spatial comparison. While higher SM in humid regions often implies greater variability, which poses a significant challenge, PSCNet effectively captures these complex dynamics, maintaining robust performance where other models falter.

\subsection{Model Evaluation Using In-situ Soil Moisture Data}

\subsubsection{Comparative Evaluation of Models at Overall and Network Scales}
\begin{figure}
	\centering
	\includegraphics[width=0.80\textwidth]{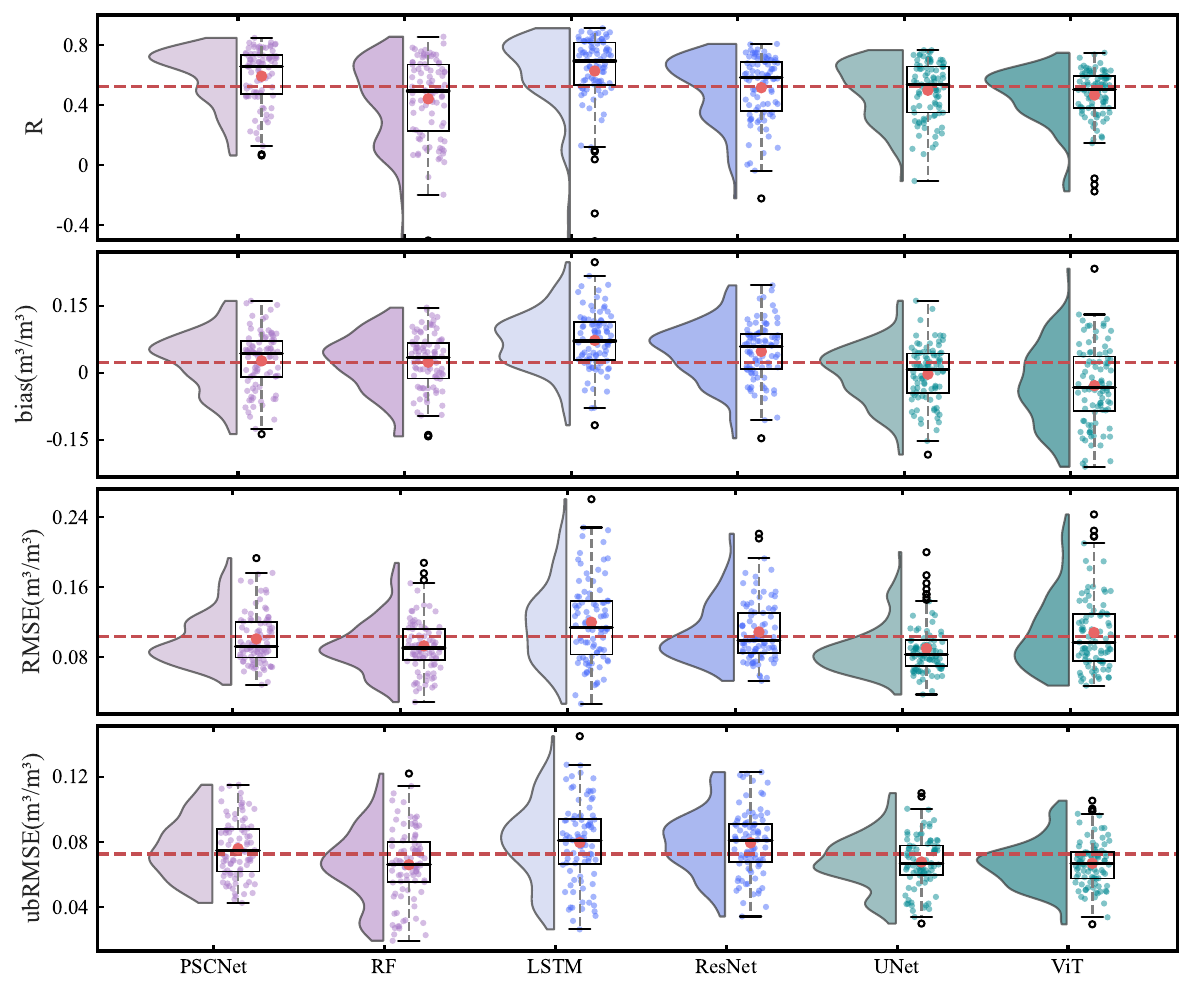}
	\caption{Cloud and rain plots of statistical metrics comparing the different downscaling models using in-situ site data (split violin plots with density distributions on the left and scatter points with box plots on the right, red dashed lines represent overall mean values, and red dots represent each method's mean performance in the dataset).}
	\label{FIG:5}
\end{figure}

To further evaluate the performance and generalization capability of the downscaling models, we directly compare their 3-hour downscaled results against in-situ SM observations from five networks on the TP (Maqu, Naqu, Ali, Shiquanhe, and CTP) during the thawing season. The comprehensive results are presented visually through raincloud plots in Fig.~\ref{FIG:5} and detailed numerically in Table~\ref{tbl3}.

\begin{table}[ht]
\caption{Statistical results of the downscaled products from different models against in-situ network data (R, bias, RMSE, and ubRMSE). The top three performers for each metric are highlighted in shades of green from dark to light.}
\label{tbl3}
\centering
\scalebox{0.8}{
\begin{tabular}{llcccccc}
\toprule
\textbf{Products} & \textbf{Metrics} & \textbf{PSCNet} & \textbf{LSTM} & \textbf{ResNet} & \textbf{UNet} & \textbf{ViT} & \textbf{RF} \\
\midrule
\multirow{4}{*}{Maqu} & R & \cellcolor[HTML]{e2efd9}0.380 & \cellcolor[HTML]{78A55C}0.419 & 0.354 & 0.347 & \cellcolor[HTML]{a8d08d}0.391 & 0.221  \\
& bias & \cellcolor[HTML]{a8d08d}-0.003 & 0.079 & 0.025 & \cellcolor[HTML]{e2efd9}-0.011 & -0.066 & \cellcolor[HTML]{78A55C}0.002 \\
& RMSE & \cellcolor[HTML]{a8d08d}0.105 & 0.138 & 0.109 & \cellcolor[HTML]{78A55C}0.103 & 0.118 & \cellcolor[HTML]{a8d08d}0.105 \\
& ubRMSE & \cellcolor[HTML]{e2efd9}0.075 & 0.091 & 0.077 & \cellcolor[HTML]{78A55C}0.070 & 0.076 &  \cellcolor[HTML]{a8d08d}0.073 \\
\midrule
\multirow{4}{*}{Naqu} & R & \cellcolor[HTML]{e2efd9}0.690 & \cellcolor[HTML]{78A55C}0.790 & 0.652 & 0.634 & 0.586 & \cellcolor[HTML]{a8d08d}0.772 \\
& bias & \cellcolor[HTML]{e2efd9}0.071 & 0.090 & 0.091 & \cellcolor[HTML]{78A55C}0.023 & \cellcolor[HTML]{a8d08d}0.025 & 0.077 \\
& RMSE & 0.118 & 0.121 & 0.131 & \cellcolor[HTML]{a8d08d}0.078 & \cellcolor[HTML]{78A55C}0.073 & \cellcolor[HTML]{e2efd9}0.101 \\
& ubRMSE & 0.091 & 0.078 & 0.093 & \cellcolor[HTML]{e2efd9}0.071 & \cellcolor[HTML]{a8d08d}0.066 & \cellcolor[HTML]{78A55C}0.060 \\
\midrule
\multirow{4}{*}{Ali} & R & \cellcolor[HTML]{78A55C}0.725 & \cellcolor[HTML]{a8d08d}0.620 & \cellcolor[HTML]{e2efd9}0.549 & 0.474 & 0.535 & 0.378 \\
& bias & 0.018 & \cellcolor[HTML]{a8d08d}0.012 & 0.031 & \cellcolor[HTML]{78A55C}-0.007 & 0.018 & \cellcolor[HTML]{e2efd9}0.015 \\
& RMSE & \cellcolor[HTML]{78A55C}0.048 & \cellcolor[HTML]{a8d08d}0.051 & 0.059 & \cellcolor[HTML]{e2efd9}0.052 & 0.052 & 0.056 \\
& ubRMSE & \cellcolor[HTML]{78A55C}0.045 & \cellcolor[HTML]{e2efd9}0.049 & 0.050 & 0.051 & \cellcolor[HTML]{a8d08d}0.049 & 0.054 \\
\midrule
\multirow{4}{*}{Shiquanhe} & R & \cellcolor[HTML]{a8d08d}0.500 & \cellcolor[HTML]{78A55C}0.534 & 0.462 & \cellcolor[HTML]{e2efd9}0.469 & 0.417 & 0.453 \\
& bias & 0.048 & \cellcolor[HTML]{a8d08d}0.032 & 0.072 & \cellcolor[HTML]{e2efd9}0.034 & 0.098 & \cellcolor[HTML]{78A55C}0.024 \\
& RMSE & 0.084 & \cellcolor[HTML]{a8d08d}0.059 & 0.098 & \cellcolor[HTML]{e2efd9}0.060 & 0.117 & \cellcolor[HTML]{78A55C}0.048 \\
& ubRMSE & 0.060 & \cellcolor[HTML]{a8d08d}0.042 & 0.060 & \cellcolor[HTML]{e2efd9}0.043 & 0.059 & \cellcolor[HTML]{78A55C}0.031 \\
\midrule
\multirow{4}{*}{CTP} & R & \cellcolor[HTML]{a8d08d}0.726 & \cellcolor[HTML]{78A55C}0.782 & 0.672 & 0.632 & 0.575 & \cellcolor[HTML]{e2efd9}0.699 \\
& bias & \cellcolor[HTML]{78A55C}0.011 & 0.061 & 0.037 & \cellcolor[HTML]{a8d08d}-0.023 & -0.056 & \cellcolor[HTML]{e2efd9}0.028 \\
& RMSE & \cellcolor[HTML]{e2efd9}0.102 & 0.118 & 0.110 & \cellcolor[HTML]{a8d08d}0.101 & 0.118 & \cellcolor[HTML]{78A55C}0.098 \\
& ubRMSE & 0.081 & 0.084 & 0.085 & \cellcolor[HTML]{e2efd9}0.072 & \cellcolor[HTML]{78A55C}0.069 & \cellcolor[HTML]{a8d08d}0.072 \\
\bottomrule
\end{tabular}
}
\end{table}

The raincloud plots in Fig.~\ref{FIG:5} provide a comprehensive overview of the statistical distribution for each model's performance against in-situ data. Among the models, LSTM exhibits the highest median value and an excellent overall distribution for the R metric against in-situ observations; however, it shows clear disadvantages in other metrics. In contrast, the PSCNet model integrates the advantages of temporal and spatial models, not only maintaining a temporal correlation comparable to that of LSTM but also exhibiting strong performance in error metrics, with RMSE and ubRMSE distributions significantly lower than LSTM's. Furthermore, the UNet, ViT, and RF models demonstrate a strong capability for error minimization, highlighting their effectiveness in leveraging spatial context for error reduction.

The statistical results of the downscaled products from different models against the in-situ sites are detailed in Table~\ref{tbl3}. The overall results are largely consistent with the findings from the raincloud plots. The LSTM model excels in temporal correlation, but its strong performance on other metrics is largely confined to the Ali and Shiquanhe networks. Conversely, models such as RF, UNet, and ViT often demonstrate robust performance in the bias, RMSE, and ubRMSE metrics. While our proposed PSCNet does not always emerge as the top performer in every network, it consistently maintains competitive and stable results across all sites.

\begin{figure}
	\centering
	\includegraphics[width=1\textwidth]{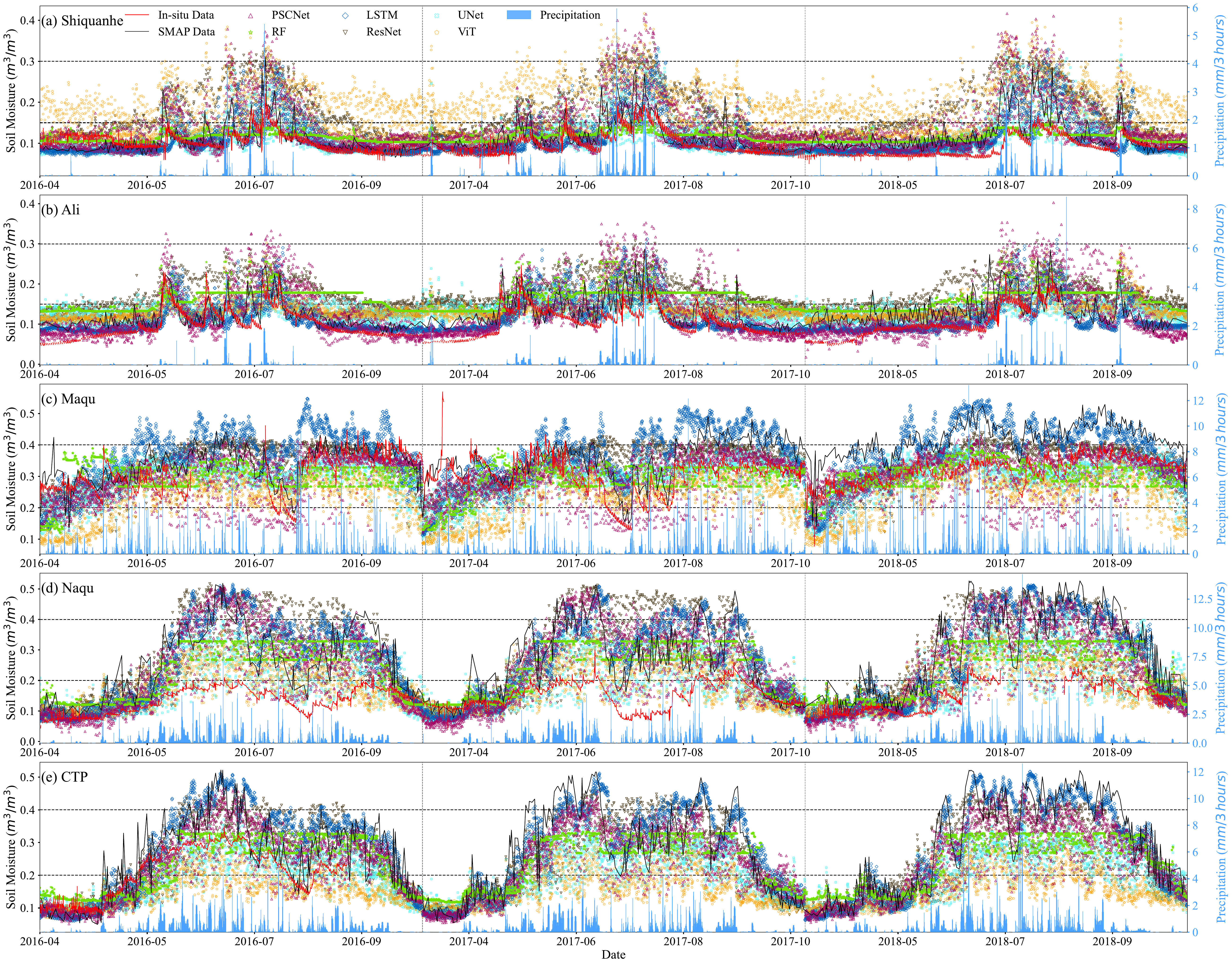}
	\caption{Time series of SM from in-situ networks, SMAP satellite products, and downscaled products, and precipitation data, for (a) Shiquanhe; (b) Ali; (c) Maqu; (d) Naqu; and (e) CTP across five networks spanning the freeze-thaw season of 2016.4-2018.12.}
	\label{FIG:9}
\end{figure}

The preceding analysis has detailed the statistical performance of the various downscaling models against in-situ station data. However, this performance is not entirely consistent with the models' fitting capabilities against the original SMAP microwave satellite data. This discrepancy stems from the inherent systematic measurement bias between satellite-based SMAP data and ground-based in-situ observations. To further investigate these temporal behaviors, we utilize temporal dynamics visualization to examine the patterns of the downscaled data across different networks. Fig.~\ref{FIG:9} shows the temporal dynamics plots constructed by sampling and averaging stations within each network, displaying three data types: in-situ observations, SMAP satellite data, and downscaled products. SMAP satellite data have a temporal frequency of 12 hours, while in-situ observations and downscaled products have a temporal frequency of 3 hours. The original SMAP microwave products exhibit varying degrees of overestimation compared to in-situ measurements across different networks, with this issue being particularly pronounced in the Naqu network. Moreover, SMAP products show substantial instability in their temporal dynamics, particularly during the rainy season with high precipitation.

A comparison of the downscaled products with the in-situ SM data reveals several key points: 1) The original SMAP microwave data exhibit a significant bias against the in-situ network measurements. This bias is particularly pronounced in networks with higher precipitation (e.g., Maqu, Naqu, and CTP) compared to the more arid networks (Shiquanhe and Ali). 2) Consequently, the SM estimates from the downscaled products also show a general overestimation relative to the in-situ data, as their training target is the biased SMAP data itself, rather than the in-situ observations.

Compared with the original SMAP satellite product sequence, the downscaled results from the various models show distinct characteristics in their temporal fidelity. LSTM and PSCNet effectively track the temporal dynamics of the SMAP sequence, with their performance being particularly strong in arid networks. In high-precipitation networks, however, their behaviors differ during the rainy season: PSCNet is more responsive to the magnitude of precipitation-induced changes, which can lead to greater prediction variability, whereas LSTM provides smoother temporal tracking, albeit with a slight overestimation. The other baseline models struggle to a greater extent in capturing these dynamics. The RF model provides overly smoothed estimates with insufficient sensitivity to sharp fluctuations. Similarly, ResNet and UNet fail to capture the full range of temporal variations, and this issue is most severe with ViT, whose predictions are erratic and substantially deviate from the observed sequence.

Overall, the systematic bias present between the original SMAP data and in-situ observations is propagated into all downscaled products through the training process, becoming most evident during the rainy season. Among the baseline models, LSTM demonstrates excellent temporal capture capabilities, while others such as RF, UNet, and ResNet are particularly effective at reducing errors and bias, our proposed PSCNet uniquely synthesizes these capabilities, delivering both robust temporal reconstruction and consistently low errors.

\subsubsection{Evaluation of the Proposed PSCNet Model at Individual Sites}
The network-level metrics offer a valuable macro-level overview. However, they inherently average out and obscure the granular, site-specific behaviors of the model. Therefore, a site-level validation is conducted here to provide a more nuanced and robust assessment of PSCNet's performance under diverse local conditions.

  \begin{figure}
    \centering
    \includegraphics[width=1\textwidth]{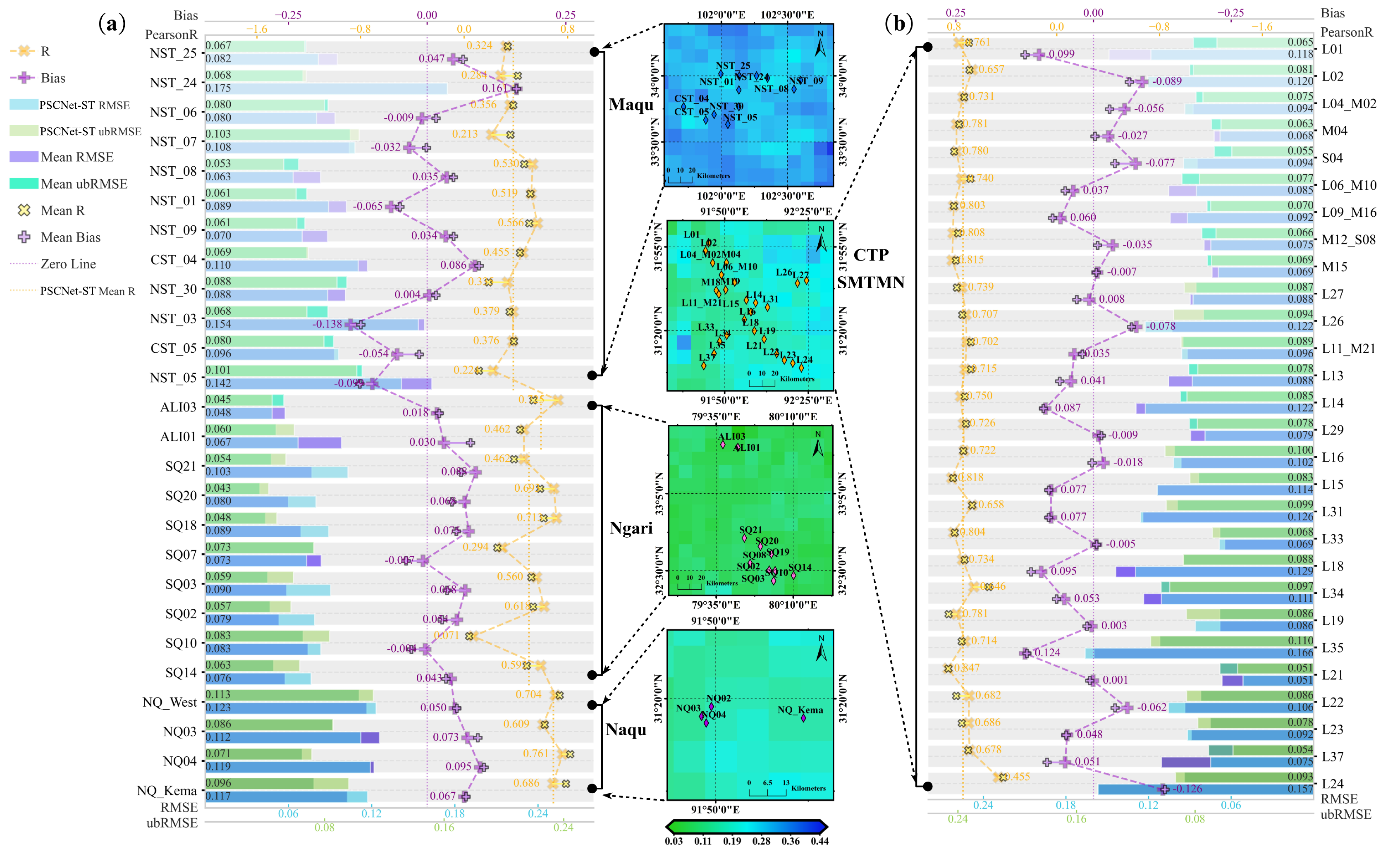}
    \caption{Validation of representative sites within the network using the PSCNet model. Panels (a) and (b) display site-specific error bar plots. (a) presents sites from the Maqu, Ngari, and Naqu networks, while (b) is dedicated to the CTP network. The figure's central section illustrates the annual averages of the SM products and their spatial distribution within each network. For validation metrics, RMSE and ubRMSE mean values are presented as stacked bar charts, visualizing the relative differences between PSCNet and the Mean model using varying transparency. Conversely, for R and Bias mean values, the line plots display scatter points for PSCNet, and for each site, an additional point representing the Mean model's performance is plotted. Furthermore, reference line (hereinafter referred to as the Zero Line) indicating the network-level mean R for all models is included in the line plots. All Mean model values (for both bar and line plots) and the Zero Line are computed from PSCNet, RF, LSTM, and Temporal. In these visualizations, only PSCNet's metrics are explicitly labeled.}
    \label{FIG:8}
  \end{figure}

As shown in Fig.~\ref{FIG:8}, the performance metrics exhibit distinct characteristics across the different networks. In networks with lower annual mean SM values, such as the Ali network, all models generally demonstrate better and more consistent metrics (bias, RMSE, ubRMSE). Conversely, PSCNet demonstrates superior performance in networks with higher annual mean SM values, where other models typically fail. For instance, PSCNet maintains consistent advantages in the Maqu network, which exhibits the highest mean SM values. Finally, PSCNet's consistent improvement in the R metric across a majority of sites, which appears independent of specific network characteristics, can be attributed to its learning paradigm of effectively integrating spatial context with temporal dependencies.

\subsection{Spatial Results of Soil Moisture Downscaling}

\begin{figure}
	\centering
	\includegraphics[width=1\textwidth]{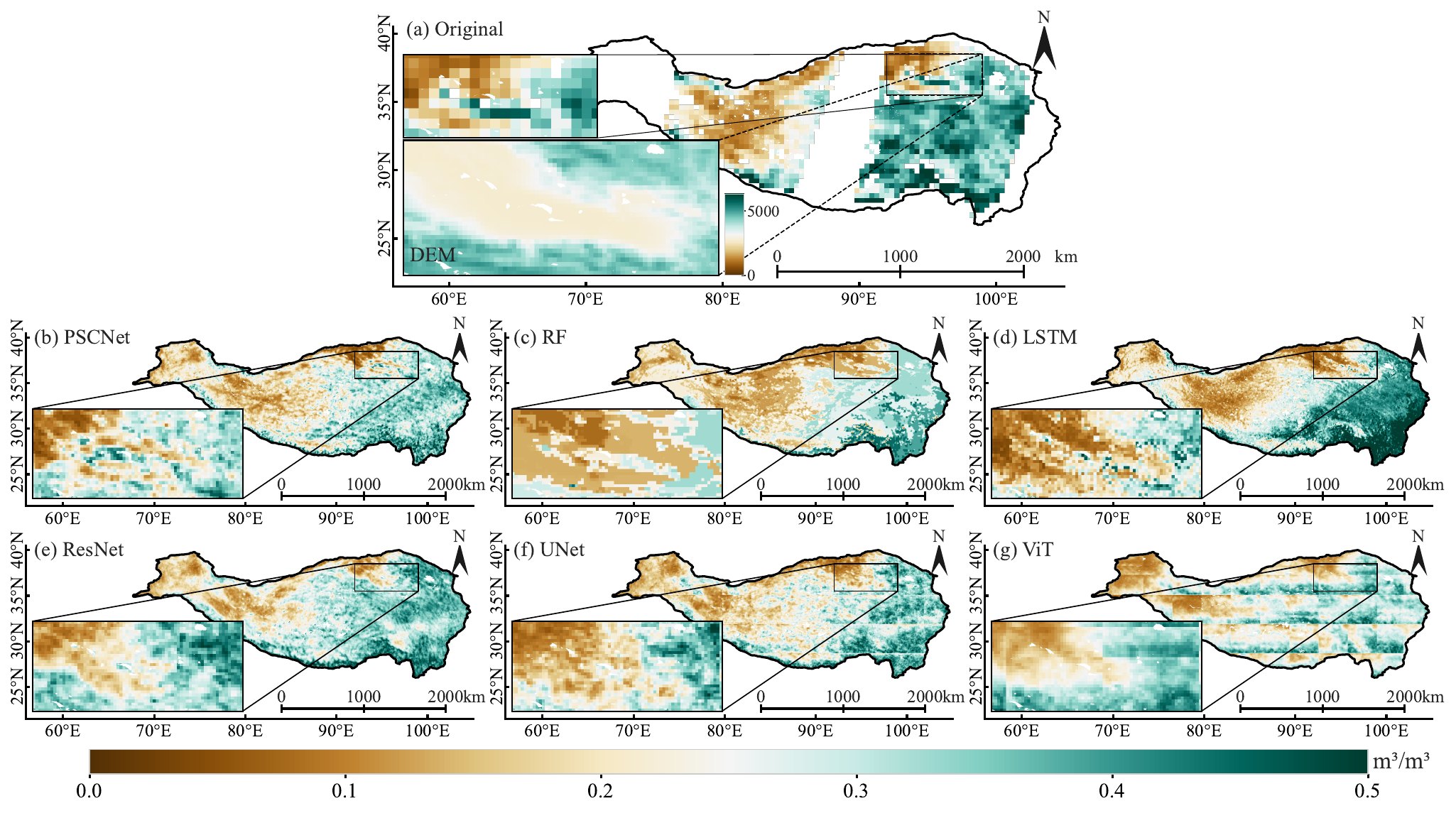}
	\caption{Visualization of SM downscaling results (open water areas masked, 2018-09-04 06:00 UTC). The Qaidam Basin was selected as the detailed visualization region. (a) Original SMAP satellite SM with additional DEM map. (b)-(g) Visualization of downscaled SM results using different downscaling methods.}
	\label{FIG:10}
\end{figure}

Based on the comprehensive visualization of the entire TP, we further present enlarged maps of the Qaidam Basin. The Qaidam Basin is a vast, enclosed intermontane rift basin characterized by a dry plateau continental climate, making it one of the plateau's most sensitive and responsive regions to climate change. As shown in Fig.~\ref{FIG:10}, in the original imagery, the periphery of the Qaidam Basin displays higher SM, while the interior features a pronounced, narrow, low-SM valley.

Visual assessment of the downscaled products will be conducted based on two criteria: (1) consistency with SMAP microwave satellite products; and (2) image smoothness and texture features. It should be noted that image smoothness and texture richness represent subjective visual comparisons and do not directly indicate reconstruction accuracy.

In terms of consistency with SMAP microwave satellite products, as illustrated in Fig.~\ref{FIG:10}, the PSCNet, RF, and LSTM models demonstrate the highest visual fidelity in matching topographical contours. However, the RF and LSTM models tends to underestimate SM values, which obscures the distinct high-SM valley in the central part of the basin. The ResNet model produces moderate results; it fails to reconstruct finer details, making the basin's distinct valley nearly indiscernible. The worst performances come from the UNet and ViT models, which exhibit noticeable tiling effects across the TP landscape. Specifically within the Qaidam Basin, the UNet model aligns more closely with topographical contours, while ViT displays the poorest performance.

In terms of image smoothness and texture features, the quality of texture reconstruction should be evaluated based on the degree of restoration to the SMAP microwave satellite data.  Models with point-based training strategies, such as RF and LSTM, show good pixel-wise accuracy but produce fragmented or discontinuous spatial patterns that compromise overall texture coherence.  Models that leverage spatial context can theoretically generate more coherent results, but the final output is significantly influenced by both the model's architecture and its inference mode. For instance, both PSCNet and ResNet employ a full-image inference strategy, which effectively mitigates the edge artifacts common in patch-based methods and yields more natural image textures. However, while this strategy enables ResNet to produce texturally coherent outputs, the early integration of spatial information inherent in its architectural design concurrently leads to an over-smoothing effect that sacrifices fine-detail reconstruction. PSCNet, conversely, leverages an architecture purposefully designed to overcome this limitation, successfully preserving fine-grained details and achieving a superior balance between texture fidelity and spatial clarity.

\section{Discussion}
In the discussion section, we examine model performance from two key perspectives: 1. temporal generalization analysis across different time points to evaluate model robustness in temporal extrapolation scenarios, and 2. inter-model uncertainty analysis using the three-cornered hat (TCH) method to quantify the consistency among different models.

\subsection{Temporal Generalization Analysis}
A primary challenge in this study stems from the temporal discrepancy between the training data and the downscaling target.  The training dataset, derived from SMAP satellite products, is limited to observations at approximately 6:00 and 18:00 UTC.  The objective, however, is to reconstruct data across the entire diurnal cycle at a 3-hour resolution.  This temporal gap raises critical concerns about the models' generalization capability when extrapolating to untrained timestamps.  Specifically, we hypothesize that the reconstruction performance may degrade as the temporal distance from the two anchor training times increases, leading to larger biases.  Therefore, this section aims to quantitatively assess this temporal generalization error to evaluate the robustness and practical applicability of each model.

In this section, we assess the feasibility of the temporal downscaling task by evaluating the performance differences among various models against in-situ station data. Specifically, we aggregate in-situ station data from multiple observation networks, partitioning daily measurements into eight discrete timestamps. We first examine the performance of different models across multiple statistical metrics at each temporal interval through comprehensive visualization analysis. Subsequently, we define temporal generalization error formulas to quantitatively assess the overall generalization capability of each model across non-training timestamps.

\begin{figure}
	\centering
	\includegraphics[width=1\textwidth]{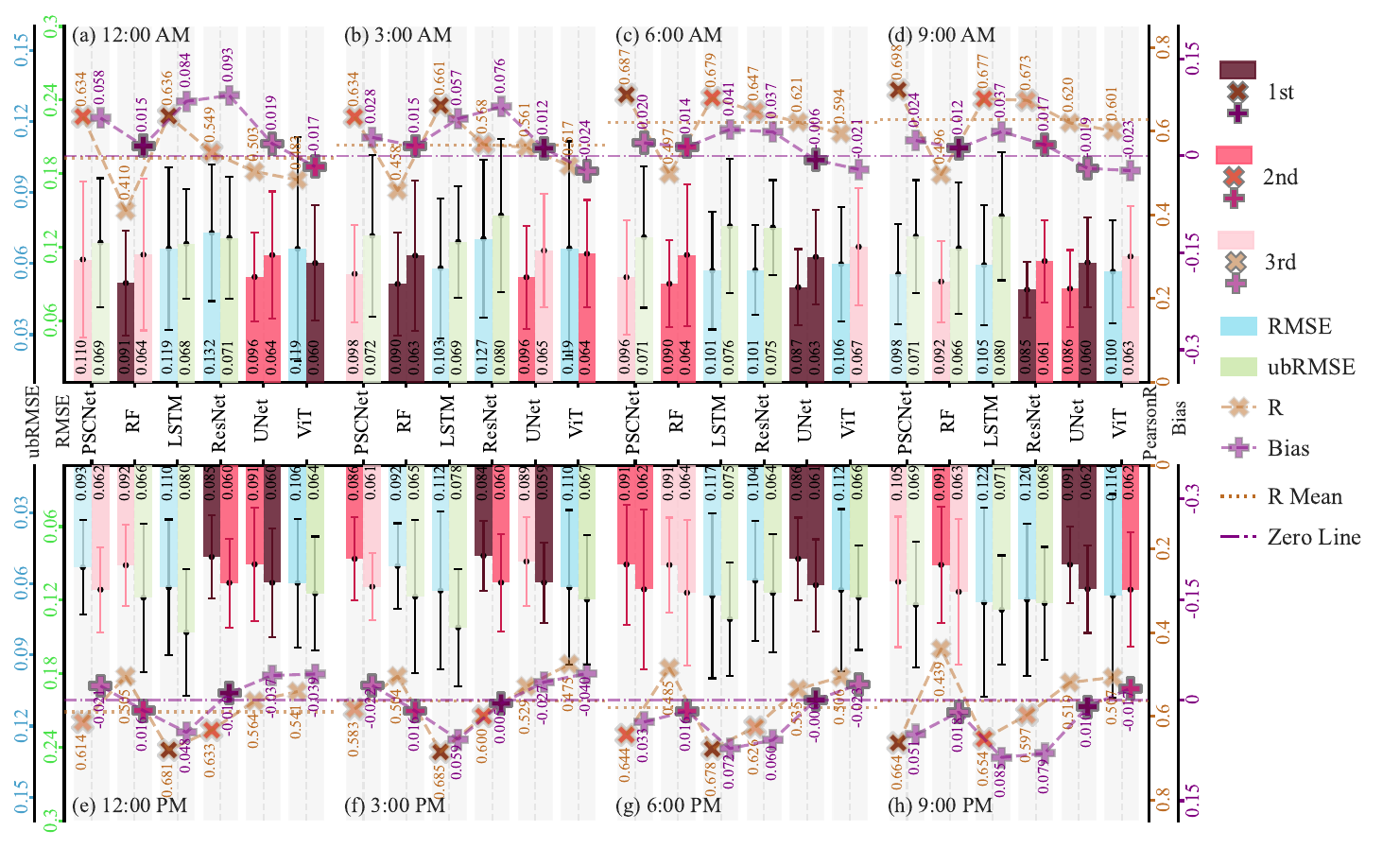}
	\caption{Plot of statistical metrics for comparisons between in-situ measurements and downscaled products across eight timestamps (0, 3, 6, 9, 12, 15, 18, 21) UTC with error bars representing the 25th and 75th percentiles of the data distribution.}
	\label{FIG:12}
\end{figure}

\begin{table}[ht]
  \caption{Relative Generalization Errors ($\overline{RE}$) for all methods across temporal intervals without training data (excluding 6:00 and 18:00 UTC), calculated according to Equation~\ref{sec:relative generalization error index}. Positive values indicate better temporal generalization capability compared to training timestamps, while negative values indicate performance degradation.}
  \label{tbl5}
  \centering
  \scriptsize
  \setlength{\tabcolsep}{2pt}
  \begin{tabular}{l|cccccc}
    \toprule
    \textbf{Metric} & \textbf{PSCNet} & \textbf{LSTM} & \textbf{ResNet} & \textbf{UNet} & \textbf{ViT} & \textbf{RF} \\
    \midrule
    $\overline{RE}_{ubRMSE}$ (\%) & -1.052 & 1.018 & 4.273 & -0.150 & \textbf{4.518} & -2.706 \\
    $\overline{RE}_{R}$ (\%) & -4.167 & \textbf{-1.937} & -5.176 & -5.007 & -5.328 & -5.153 \\
    \bottomrule
  \end{tabular}
\end{table}

\begin{table}[ht]
  \caption{Detailed Relative Generalization Errors ($\mathrm{RE}$, in \%) for all models. This table presents the $\mathrm{RE}$ at individual non-training timestamps and the overall mean error ($\overline{\mathrm{RE}}$), calculated according to Equation~\ref{sec:relative generalization error index}. Bold values indicate the worst-case degradation (minimum $\mathrm{RE}$) for each model and metric.}
  \label{tbl:tbl5}
  \centering
  \scriptsize
  \setlength{\tabcolsep}{3.5pt}
  \begin{tabular}{l|ccccccc|ccccccc}
    \toprule
    \multirow{2}{*}{\textbf{Model}} & \multicolumn{7}{c|}{\textbf{$\mathrm{RE}_{R}$ (\%)}} & \multicolumn{7}{c}{\textbf{$\mathrm{RE}_{\mathrm{ubRMSE}}$ (\%)}} \\
    \cmidrule(lr){2-8} \cmidrule(lr){9-15}
    & 00:00 & 03:00 & 09:00 & 12:00 & 15:00 & 21:00 & \textbf{$\overline{\mathrm{RE}}$} & 00:00 & 03:00 & 09:00 & 12:00 & 15:00 & 21:00 & \textbf{$\overline{\mathrm{RE}}$} \\
    \midrule
    PSCNet & -4.69  & -4.79  & 4.91  & -7.71  & \textbf{-12.45} & -0.27  & -4.17  & -3.17  & \textbf{-7.42} & -7.03  & 6.34   & 8.22   & 3.25   & -1.05 \\
    RF     & \textbf{-16.59} & -6.73  & 0.94  & 2.74   & 2.68   & -10.62 & -4.60  & -0.04  & 0.58   & \textbf{-4.35}  & -3.39  & -2.74  & 0.50   & -1.57 \\
    LSTM   & \textbf{-6.31}  & -2.63  & -0.28 & 0.27   & 0.96   & -3.64  & -1.94  & 9.31   & 8.20   & -6.22  & \textbf{-6.83} & -4.12  & 5.76   & 1.02 \\
    ResNet & \textbf{-13.71} & -10.73 & 5.82  & -0.48  & -5.75  & -6.20  & -5.18  & -1.87  & \textbf{-15.42} & 12.25  & 14.28  & 14.46  & 1.94   & 4.27 \\
    UNet   & \textbf{-12.99} & -2.96  & 7.17  & -2.51  & -8.55  & -10.19 & -5.01  & -3.08  & \textbf{-6.08} & 2.09   & 3.44   & 3.59   & -0.86  & -0.15 \\
    ViT    & -12.11 & -6.03  & 9.19  & -1.63  & \textbf{-13.56} & -7.82  & -5.33  & 9.36   & 3.49   & 5.19   & 3.56   & \textbf{-0.41} & 5.93   & 4.52 \\
    \bottomrule
  \end{tabular}
\end{table}

The visualization analysis reveals distinct temporal generalization patterns across different model architectures. In Fig.~\ref{FIG:12}, the sequential model LSTM achieves the highest R values in more than half of the timestamps. However, it exhibits weaker performance in terms of bias, RMSE, and ubRMSE. In contrast, the spatial models UNet and ResNet, along with the point-based model RF, demonstrate strong performance in terms of bias, RMSE, and ubRMSE, consistently outperforming the sequential model on these metrics during most time periods. Building on these complementary strengths, PSCNet, with its spatio-temporal architecture, successfully integrates the advantages of both sequential and spatial models. Specifically, it maintains a significant advantage in R, achieving the highest values at three timestamps (6:00, 9:00, and 21:00) and ranking within the top three at the remaining timestamps. Furthermore, it outperforms the sequential model in terms of bias, RMSE, and ubRMSE across most networks.

The quantitative results in Table~\ref{tbl:tbl5} provide a detailed assessment of the temporal generalization capability of each model. In terms of ($\overline{\mathrm{RE}}_{R}$), models that incorporate sequential modeling, namely LSTM and PSCNet, demonstrate less performance degradation than their counterparts, with the average degradation across all models being approximately 5\%. In contrast, for $\overline{\mathrm{RE}}_{\mathrm{ubRMSE}}$, the spatially explicit models perform exceptionally well, with ViT and ResNet yielding the best results (4.52\% and 4.27\%, respectively). The average degradation for the other models is approximately 1\%, suggesting that the ubRMSE metric is less sensitive to performance decay from temporal extrapolation. A closer look at timestamp-specific errors reveals that for the $\mathrm{RE}_{R}$ metric, the most significant degradation for a majority of models clusters at 00:00. This is likely attributable to two primary factors: its maximal temporal distance from the training anchors (06:00 and 18:00 UTC) and the discrepancy between instantaneous nocturnal conditions and daily-updated auxiliary variables. In contrast, the temporal distribution of degradation for $\mathrm{RE}_{\mathrm{ubRMSE}}$ does not exhibit such a distinct pattern. Overall, these results validate the feasibility of using the integration of validated high-spatio-temporal resolution reanalysis variables for spatio-temporal downscaling on the TP, as the average temporal generalization error remains within 6\% for the R metric and 2\% for the ubRMSE metric.

\subsection{Uncertainty Analysis}
\begin{figure}
	\centering
	\includegraphics[width=1\textwidth]{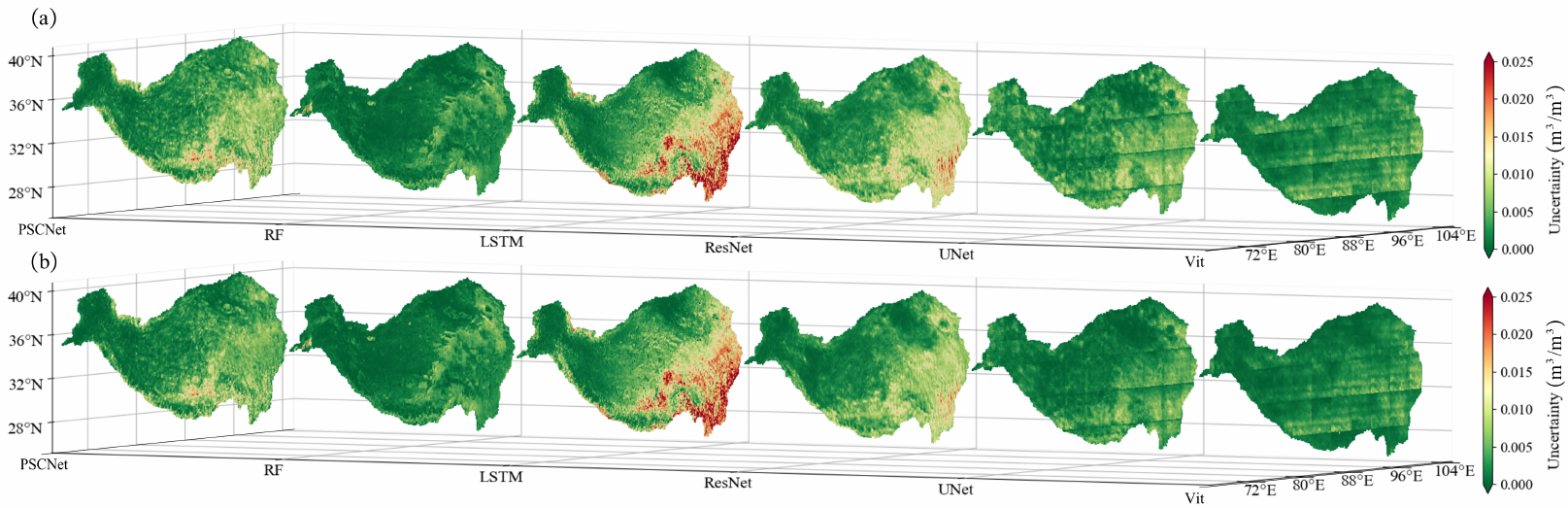}
	\caption{Uncertainties calculated by the TCH methods, (a) (c), (b) (d) Plots of the spatial distribution of the TCH uncertainties of each method after a sampling frequency of 3H and daily averaging, respectively.}
	\label{FIG:11}
\end{figure}

Due to the scarcity of in-situ SM measurement data on the TP, comprehensively assessing the reconstruction performance of various models presents a significant challenge. One approach is to use the TCH method to examine inter-model consistency when only limited in-situ observations are available~\citep{P_Tavella_1994}. Notably, the TCH method quantifies inter-model uncertainty rather than absolute accuracy, with lower uncertainty indicating proximity to the ensemble mean but not necessarily superior performance. For this uncertainty analysis, we selected PSCNet-series models and point-based models for comparison. Fig.~\ref{FIG:11} demonstrates the uncertainties among the five models calculated using the TCH method.

The TCH analysis reveals that the spatial distribution of uncertainty across various models exhibits a consistent primary pattern: uncertainty gradually increases from the Pamir Plateau, through the Karakoram Range and Northern TP, to the Hengduan Mountains, Tanggula Mountains, and the eastern TP. This spatial pattern directly corresponds to the gradient of SM values, indicating that greater inter-model disagreement occurs in regions with higher SM. Beyond this overarching trend, the varied spatial distributions also highlight the region-specific strengths and weaknesses of different modeling approaches. For instance, models such as PSCNet, LSTM, and ResNet, which demonstrated superior performance in high-SM regions during the spatial validation (Fig.~\ref{FIG:7}), paradoxically exhibit far greater uncertainty in these areas compared to other models. This is particularly the case for the LSTM model, which is prone to higher uncertainty in regions with dramatic elevation changes, such as the Gangdise, Hengduan, and Bayan Har Mountains. Furthermore, these inter-model discrepancies are sensitive to the temporal scale, becoming more pronounced at the 3-hourly resolution (Fig.~\ref{FIG:11} (a)) than at the daily scale (Fig.~\ref{FIG:11} (b)), which is particularly evident in the UNet and ViT products. In contrast, the RF model demonstrates greater stability, maintaining favorable uncertainty performance across both resolutions.

\section{Conclusions and future work}
Focusing on high-resolution spatio-temporal downscaling, this study investigates two modeling paradigms: spatially explicit models and point-based models, to reconstruct SM data over the TP at 10-km spatial and 3-hour temporal resolution from 2016 to 2018. Our results indicate that the proposed PSCNet spatio-temporal model outperforms various spatial models in satellite product validation, in-situ site validation, and other validation experiments, demonstrating superior accuracy and stability. Additionally, the model shows robust performance in predicting previously unseen time points. In temporal dynamics and visualization validation, PSCNet demonstrates excellent temporal sensitivity and vivid spatial details. These results empirically validate the enhanced generalization capability of our model, emphasizing its potential for application in increasingly complex environments.

The observed performance improvements can be attributed to the model’s robust spatio-temporal modeling of SM. By leveraging multi-source auxiliary variables and multi-frequency temporal data, the model effectively captures the nonlinear mapping relationships of SM. This representation enables the model to handle complex scenarios driven by rainfall variability and adapt to dynamic meteorological conditions. These findings underscore the potential of spatio-temporal modeling in advancing the understanding of SM dynamics.

Although this study successfully achieves temporal upsampling to a 3-hour resolution, there is still considerable room for improvement in spatial reconstruction at the 10-km scale. Future work will focus on developing enhanced spatio-temporal reconstruction frameworks. Current limitations include the lack of systematic variable screening that incorporates surface auxiliary features, as well as limited model evaluation across different land cover types, human activities, and climatic conditions. These issues warrant further investigation in subsequent research.







\appendix
\section{Appendix}
\subsection{Statistical indicators}\label{sec:statistical_indicators}
The specific expressions of the four evaluation indicators used in the study are as follows:
\setcounter{equation}{0}
\renewcommand{\theequation}{A.\arabic{equation}}
\begin{equation}
  \mathrm{RMSE}=\sqrt{\frac{\sum_{i=1}^N{\mathrm{}}\left( x_i-y_i \right) ^2}{N}}
\end{equation}
\begin{equation}
  \mathrm{ubRMSE}=\sqrt{\frac{\sum_{i=1}^N{\mathrm{}}\left[ \left( x_i-\bar{x} \right) -\left( y_i-\bar{y} \right) \right] ^2}{N}}
\end{equation}
\begin{equation}
  \mathrm{R} = \frac{\sum_{i=1}^N \left( \left( x_i - \bar{x} \right) \cdot \left( y_i - \bar{y} \right) \right)}{\sqrt{\sum_{i=1}^N \left( x_i - \bar{x} \right)^2 \cdot \sum_{i=1}^N \left( y_i - \bar{y} \right)^2}}
\end{equation}

\begin{equation}
  \mathrm{bias}=\frac{1}{N}\sum_{i=1}^N{\mathrm{}}\left( x_i-y_i \right) 
\end{equation}
$x_i$ and $y_i$ respectively represent the actual observed SM data and the predicted SM data corresponding to the observation time $i$, while $\bar{x}$ and $\bar{y}$ respectively represent the overall mean of the actual and predicted SM data during the observation period.

\subsection{Relative generalization error index}\label{sec:relative generalization error index}
In this paper, the relative generalization error of various indicators is calculated as follows
\begin{equation}
  \bar{X}_{\mathrm{metric}}=\frac{X_{\mathrm{metric}}^{6}+X_{\mathrm{metric}}^{18}}{2}
\end{equation}
In the formula, $X_{\mathrm{metric}}^i$ represents the value of the metric at time $i$, and $\bar{X}_{\mathrm{metric}}$ is the mean value of the metric at the time points present in the training dataset, serving as a baseline for validation. This includes metrics such as $\mathrm{R}$ and $\mathrm{ubRMSE}$. Following this, the Relative Error ($\mathrm{RE}$) at an individual non-training timestamp $j$ is calculated. The formulas are defined such that a positive value indicates improved performance relative to the baseline:
\begin{equation}
  \mathrm{RE}_{\mathrm{ubRMSE}}(j) = \frac{\bar{X}_{\mathrm{ubRMSE}} - X_{\mathrm{ubRMSE}}^j}{\bar{X}_{\mathrm{ubRMSE}}}
\end{equation}
\begin{equation}
  \mathrm{RE}_R(j) = \frac{X_R^j - \bar{X}_R}{\bar{X}_R}
\end{equation}

Finally, the mean Relative Generalization Error ($\overline{\mathrm{RE}}$) is computed by averaging the individual RE values across the $N=6$ non-training timestamps $T = \{0, 3, 9, 12, 15, 21\}$:
\begin{equation}
  \overline{\mathrm{RE}}_{\mathrm{metric}} = \frac{1}{N} \sum_{j \in T} \mathrm{RE}_{\mathrm{metric}}(j)
\end{equation}

\printcredits
\section*{Declaration of competing interest}
The authors declare that they have no known financial or personal relationships that could be perceived as competing interests, which might have influenced the research presented in this paper.

\section*{Acknowledgments}
This research was supported in part by the National Natural Science Foundation of China (42271324, 42201406), the Natural Science Foundation of Jiangsu Province, China (BK20221506), the Natural Science Foundation for Colleges and Universities of Jiangsu Province under Grant 22KJB570006, and the Qing Lan Project of Jiangsu Province of China.
\bibliographystyle{cas-model2-names.bst}

\bibliography{cas-refs.bib}



\end{document}